\documentclass[letterpaper, 10 pt, conference]{ieeeconf}  

\IEEEoverridecommandlockouts                              

\title{\LARGE \bf
Learning Stixel-based Instance Segmentation
}

\author{Monty Santarossa$^{1}$, Lukas Schneider$^{2}$, Claudius Zelenka$^{1}$, Lars Schmarje$^{1}$, Reinhard Koch$^{1}$ and Uwe Franke$^{3}$
\thanks{$^{1}$These authors are with Multimedia Information Processing,
        Kiel University, 24118 Kiel, Germany
        {\tt\small \{msa, cze, las, rk\}@informatik.uni-kiel.de}}%
\thanks{$^{2}$This author is with Daimler AG R\&D, 70565 Stuttgart, Germany
        {\tt\small lukas.schneider@daimler.com}}%
\thanks{$^{3}$This author was formerly associated with Daimler AG R\&D
        {\tt\small dr.uwe.franke@gmail.com}}%
}

\usepackage[T1]{fontenc} 

\let\labelindent\relax 
\usepackage[inline]{enumitem}
\usepackage{capt-of}
\usepackage[caption=false]{subfig}

\usepackage[export]{adjustbox}
\usepackage{threeparttable}
\usepackage{comment}
\usepackage{nidanfloat}
\usepackage{cite}

\usepackage{url}

\usepackage[absolute,overlay]{textpos} 

\begin{document}

\maketitle
\thispagestyle{empty}
\pagestyle{empty}

\begin{textblock*}{0.8\paperwidth}(0.1\paperwidth,0.95\paperheight) 
   \footnotesize{© 2021 IEEE.  Personal use of this material is permitted.  Permission from IEEE must be obtained for all other uses, in any current or future media, including reprinting/republishing this material for advertising or promotional purposes, creating new collective works, for resale or redistribution to servers or lists, or reuse of any copyrighted component of this work in other works.}
\end{textblock*}

\begin{abstract}

Stixels have been successfully applied to a wide range of vision tasks in autonomous driving, recently including instance segmentation. However, due to their sparse occurrence in the image, until now Stixels seldomly served as input for Deep Learning algorithms, restricting their utility for such approaches.
In this work we present StixelPointNet, a novel method to perform fast instance segmentation directly on Stixels. By regarding the Stixel representation as unstructured data similar to point clouds, architectures like PointNet are able to learn features from Stixels. We use a bounding box detector to propose candidate instances, for which the relevant Stixels are extracted from the input image. On these Stixels, a PointNet models learns binary segmentations, which we then unify throughout the whole image in a final selection step. StixelPointNet achieves state-of-the-art performance on Stixel-level, is considerably faster than pixel-based segmentation methods, and shows that with our approach the Stixel domain can be introduced to many new 3D Deep Learning tasks.

\end{abstract}

\section{Introduction}
\label{intro}

   \begin{figure*}[b]
     \subfloat[RGB with Stixel overlay\label{fig:intro1}]{%
       \includegraphics[width=0.32\linewidth]{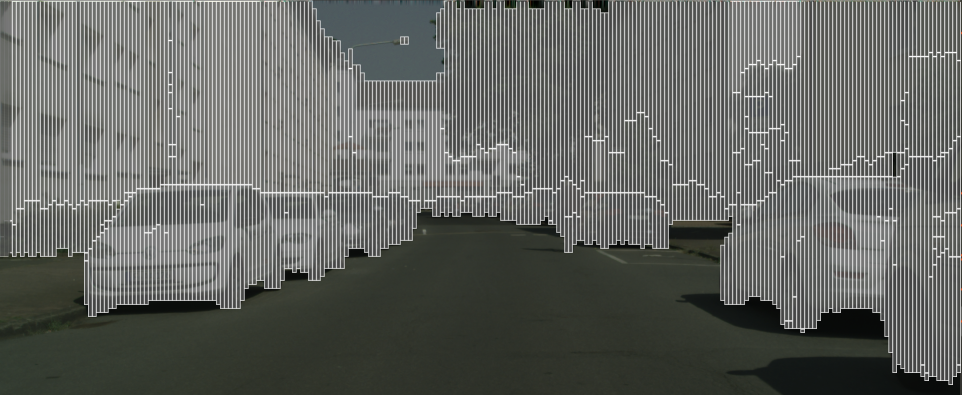}
     }
     \hfill
     \subfloat[Semantic segmentation\label{fig:intro2}]{%
        \includegraphics[width=0.32\linewidth]{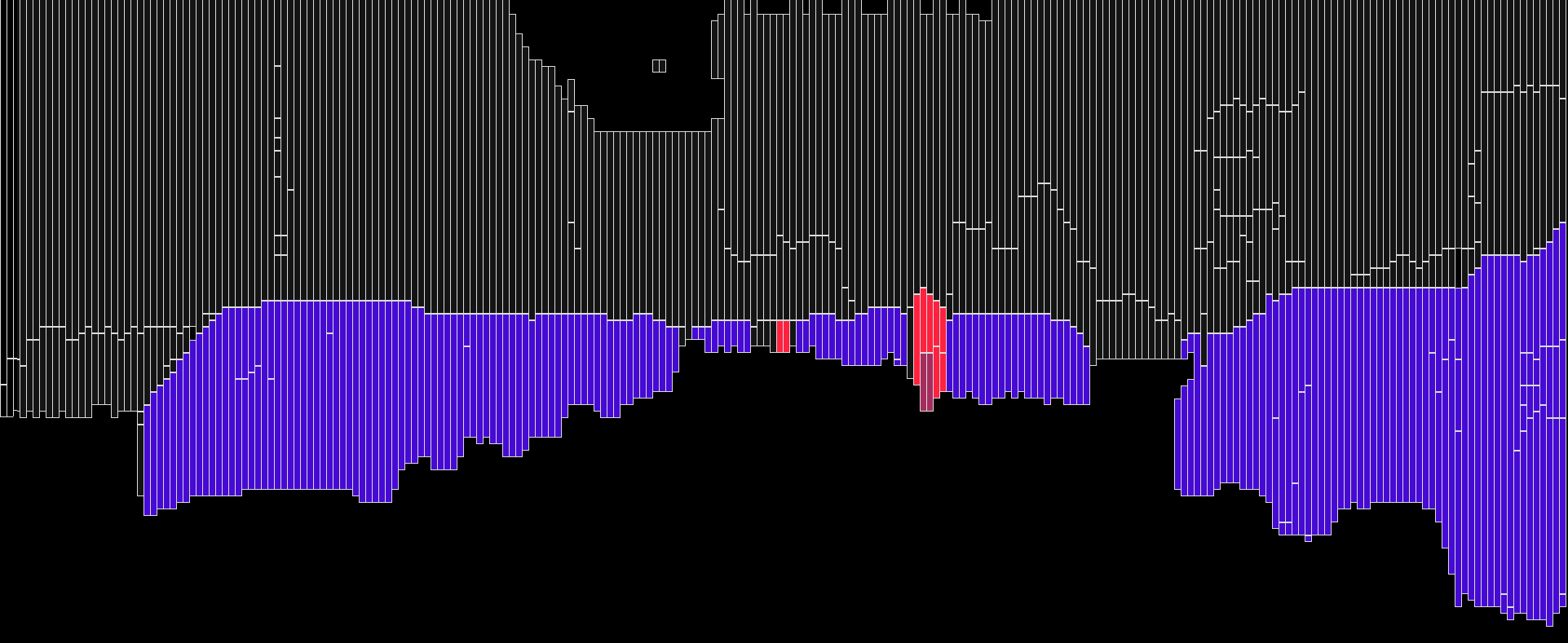}
     }
     \hfill
     \subfloat[Instance segmentation\label{fig:intro3}]{%
        \includegraphics[width=0.32\linewidth]{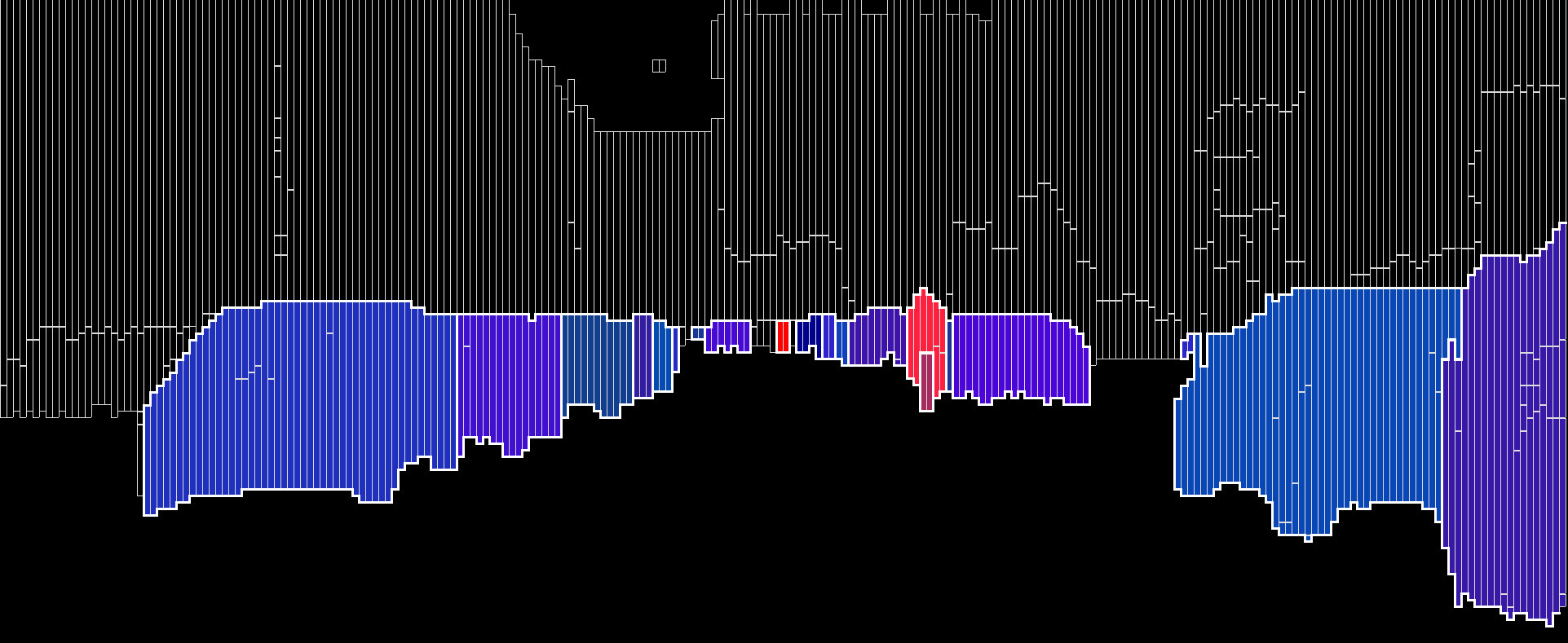}
     }
     \caption{Stixels represent objects in the world via rectangular sticks (a). While Semantic Stixels allow to associate each Stixel with a semantic class (b), instance Stixels allow to differentiate objects of the same class (c). Here, Stixels of the same color belong to the same instance}
     \label{fig:intro}
   \end{figure*}

Fully autonomous driving has proven to be a challenging, yet exciting task both in the field of computer vision and machine learning. Safe navigation on public roads is dependent on the identification of potential obstacles in real time. The Stixel World \cite{badino2009stixel,pfeiffer2011towards} as a medium-level representation of the world that is compact, yet general has become an established model for urban environments \cite{cordts2017stixel}.

\begin{figure}[t]
\centering
\includegraphics[width=0.85\linewidth]{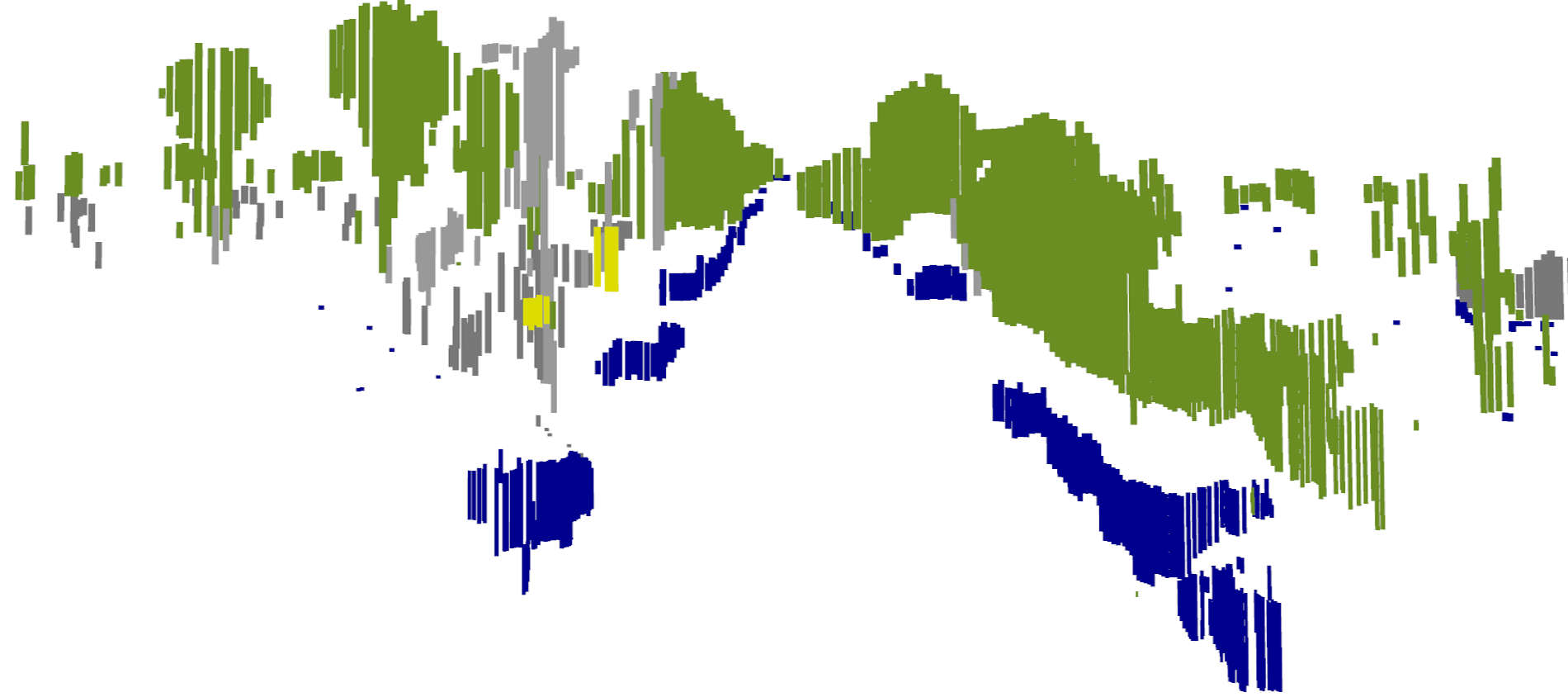}
\caption{Semantic Stixels in 3D. Colors indicate semantic classes. Due to spatial sparsity and being unstructured, Stixels can be considered complex points in a point cloud}
\label{fig:StixelUnstructured}
\end{figure}

Based on stereo disparity images \cite{hirschmuller2005accurate}, the Stixel World models the environment via a set of rectangular sticks: the Stixels. Instead of millions of pixels, each image can be represented by a few hundred Stixels, reducing both data volume and noise in depth. In the 2D image, Stixels are organized into columns as depicted in Fig \ref{fig:intro}, where their height and position can vary and irrelevant image areas are left unoccupied. In addition, Stixels possess valuable 3D information (position and shape) as illustrated in Fig. \ref{fig:StixelUnstructured}. They can also carry additional attributes like motion data \cite{pfeiffer2010efficient,pfeiffer2011modeling} or semantic labels \cite{schneider2016semantic}. Stixels have been successfully used for tasks like object detection \cite{levi2015stixelnet,benenson2012fast,li2016new} and in the machine vision system of real autonomous cars \cite{franke2013making,ziegler2014making}. Recently, first steps were taken to augment Stixels with data gained from instance segmentation \cite{hehn2019instance}.

Instance segmentation is of high relevance for the environment perception of autonomous vehicles, as it allows the differentiation of single objects with high accuracy. Such accuracy is invaluable for tasks like object tracking or movement prediction. However, performing instance segmentation in urban environments is challenging due to many occlusions and the presence of objects with different scales. While over the last years instance segmentation of pixel images has been addressed with increasingly high accuracy \cite{alhaija2017augmented,liu2018path,neven2019instance, xiong2019upsnet, peng2020deep}, it remains a costly procedure, with state-of-the-art CNN-based methods reaching less than 5 fps on Cityscapes \cite{xiong2019upsnet, peng2020deep}. 

Stixels are faster, but their sparse and unstructured properties in 2D space prove to be difficult for CNNs to account for. However, we note that these same properties in 3D space are shared by point clouds as they come e.g. from LIDAR \cite{tatoglu2012point}, inspiring this work to build upon recent successes in the point cloud domain.

This paper proposes a novel instance segmentation pipeline called \textit{StixelPointNet} that works directly on Semantic Stixels. The Stixel image is considered a complex point cloud (see Fig. \ref{fig:StixelUnstructured}), allowing to treat the 2D and semantic features of the Stixels similar to attributes like reflectance in LIDAR points. This enables point-based deep learning architectures like PointNet to learn segmentations from Stixels. StixelPointNet therefore takes advantage of the reduced data volume, allowing for real-time instance segmentation, while preserving the explicitness necessary to accurately perceive urban environments.

\section{Related Work}

\subsection{Feature Learning on Point Clouds}
A common approach to account for unordered sets of data points with varying numbers is their quantization. Especially voxel based approaches \cite{tchapmi2017segcloud,le2018pointgrid,wu20153d}, oftentimes coupled with volumetric CNNs \cite{maturana2015voxnet,huang2016point,qi2016volumetric,riegler2017octnet,dai2017scannet}, have seen extended use. However, due to the sparse nature of point clouds, quantization can lead to inefficient representation. In addition, the comparatively high computational cost of volumetric CNNs severly limits voxel size.

In 2016 Qi et al. introduced with PointNet \cite{qi2017pointnet} a novel deep learning architecture, which is able to learn directly on point clouds. Due to the use of shared MLPs (Muliti Layer Perceptrons) and a symmetric function, no quantization is required. Features are learned for each data point separately and then aggregated with a global feature learned for the entire input. As the global feature is obtained through a symmetric function (e.g. max pooling), its shape is invariant to the input order and size. Several advanced point-based architectures have since improved on PointNets generalisation abilities by capturing a better spatial context through hierarchical structures or improved feature extractor modules \cite{qi2017pointnet++,li2018pointcnn,jiang2018pointsift}. Inspired by that progress, this work regards Stixels as a complex point cloud, where each Stixel in addition to its coordinates in 3D space holds a number of additional features.

\subsection{Instance Segmentation on Point Clouds}
Building upon the accomplishments of Mask R-CNN \cite{he2017mask} in the 2D instance-segmentation domain, several similar proposal-based approaches were introduced for the task of instance segmentation of 3D point clouds. 

\textit{Region based-PointNet} \cite{yi2019gspn} follows the high-level architecture of Mask R-CNN by utilizing an object proposal module followed by a classification module, a proposal refinement module and a segmentation module. All of these use PointNet architectures. Instead of regressing bounding boxes, object proposal is achieved through a generative approach.

\textit{Frustrum PointNet} \cite{qi2018frustum} utilizes CNN-based 2D region proposals on RGBD images to create a viewing frustrum in 3D space. Pixels in the RGBD images are interpreted as points in a point cloud and filtered according to the frustrum. A PointNet model then calculates a binary segmentation, while another PointNet model predicts a 3D bounding box.

\textit{3D-BoNet} \cite{yang2019learning} avoids region proposals by introducing a novel bounding box prediction module working on point features learned by a PointNet. Two parallel models regress the bounding box coordinates and calculate a binary segmentation on the points inside the bounding box.

StixelPointNet follows the general concept of established methods so far, as regions proposals and PointNet-based binary segmentation are concerned. However, the more complex nature of Stixels compared to point clouds necessitates an adaption of those methods. The region proposals for example need to handle cases, where Stixels only partially intersect the region box.

\subsection{Instance Segmented Stixels}
\label{related:onStx}


Recently, Hehn et al. proposed {\it Instance Stixels} \cite{hehn2019instance}. By extracting instance information from the color image via a CNN, Stixels can be augmented with an instance ID. The augmentation can be applied either during Stixel generation or as a post-processing step to cluster Semantic Stixels according to the predicted instances. The former provides better results as Instance Stixels supposedly adhere better to object boundaries than Semantic Stixels. However, it is our hypothesis that the Stixel generation algorithm benefits only little from instance information, since depth alone already forces Stixels that respect object boundaries.

This hypothesis is strongly supported by the results of \cite{hehn2019instance}: Tests on Cityscapes' \cite{cordts2016cityscapes} validation dataset showed that augmentation during Stixel generation instead of clustering as a post-processing step does not improve semantic IoU (65.2 vs. 65.2) and only improves AP$_{50\%}$ by 1.6\% (21.8 vs. 20.2). At the same time, these minor improvements come at the prize of adding a costly quadratic term to the computation complexity of Stixel generation.

In contrast, our StixelPointNet works directly on the less costly Semantic Stixels. Furthermore, since StixelPointNet performs reasoning in 3D, in the future information from a variety of other sensors can be fused.



\subsection{Contributions}
This work's key contributions are:
\begin{itemize}
    \item An approach to automatically generate Stixel-level instance ground from pixel-level annotations to account for the current lack of Stixel-based datasets
    \item A novel and fast Stixel-based deep learning approach called \textit{StixelPointNet} that works on Stixels in 3D space using a PointNet model and achieves state-of-the-art instance segmentation performance on Stixel-level
    \item The implementation of several baselines for the Stixel-level dataset including established clustering methods adapted for the new Stixel domain
    \item A runtime analysis showing that StixelPointNet is real-time capable on hardware where CNN-based approaches reach less than 5 FPS
  \end{itemize}

On a general level, the achieved results show that point-based Deep Learning in the largely unexplored 3D Stixel domain is feasible, opening up a wide range of new potential tasks for Stixels, e.g. 3D bounding box estimation.

\section{Method}
\label{m:}

\begin{figure}[t]
\vspace{5mm}
\centering
\includegraphics[width=0.98\linewidth]{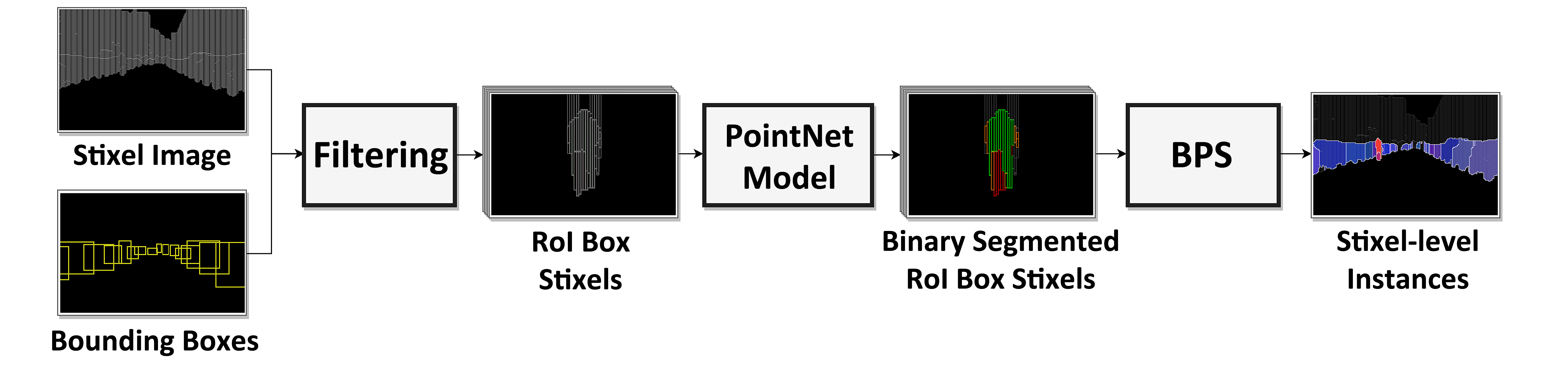}
\caption{The proposed StixelPointNet instance segmentation pipeline. The \textit{Filtering} component filters for each bounding box potential object Stixels associated to the instance the bounding box represents. The \textit{PointNet Model} predicts a binary segmentation for each bounding box, i.e. it predicts for each Stixel, whether it belongs to the instance the bounding box represents. The \textit{BPS} component (Best Prediction Selection) converts that output for all bounding boxes to one instance segmented Stixel image}
\label{fig:pipeline}
\end{figure}

Fig. \ref{fig:pipeline} shows the StixelPointNet system pipeline. It takes as input a Stixel image, i.e. a set of $N$ Semantic Stixels \cite{schneider2016semantic}, and a set of $K$ bounding boxes for that image. Semantic Stixels contain a feature vector $Stx = [x,\: y,\: z,\: w,\: h,\: u_{tl},\: v_{tl},\: u_{br},\: v_{br},\: l,\: c]$, where {\it x, y, z} define the position, {\it w} the width and {\it h} the height in 3D space, \(u_{tl}\), \(v_{tl}\), \(u_{br}\), \(v_{br}\) the rectangular position and shape in 2D image space and {\it l} and \(c\) the semantic label and its confidence respectively. Each bounding box is represented by a vector $BB = [u_{tl},\: v_{tl},\: u_{br},\: v_{br},\: l_{bb},\: c_{bb}]$, where \(u_{tl}\), \(v_{tl}\), \(u_{br}\), \(v_{br}\) describe the rectangular position and shape in 2D image space and \(l_{bb}\), \(c_{bb}\) the predicted class label and its confidence for the bounding box object. The output of the pipeline is for each Stixel an instance ID, containing the class and an instance counter. For each Stixel its instance ID is predicted through the use of the three components \textit{Filtering}, \textit{PointNet Model} and \textit{BPS} as depicted in Fig. \ref{fig:pipeline}.

\subsection{Filtering}
\label{m:filtering}

\begin{figure}[t]
     \subfloat[\label{fig:filtering0}]{%
       \includegraphics[width=0.23\linewidth]{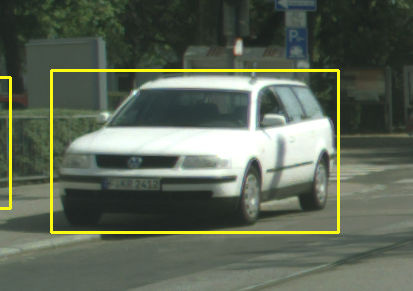}
     }
     \hfill
     \subfloat[\label{fig:filtering1}]{%
        \includegraphics[width=0.23\linewidth]{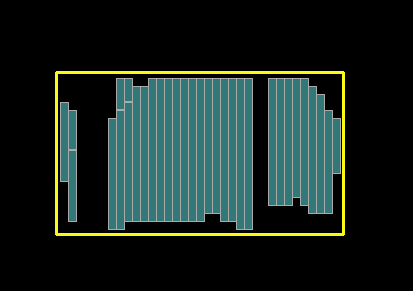}
     }
     \hfill
     \subfloat[\label{fig:filtering2}]{%
       \includegraphics[width=0.23\linewidth]{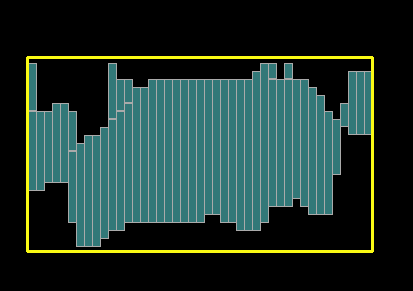}
     }
     \hfill
     \subfloat[\label{fig:filtering3}]{%
        \includegraphics[width=0.23\linewidth]{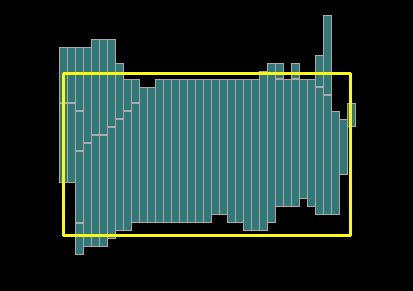}
     }
     \caption{(a) shows the SSD box in the RGB images, (b) shows the Stixels captured by this SSD box, (c) shows the Stixels captured by the RoI box with \(sc_{RoI} = 1.4\), (d) shows the Stixels captured by the RoI box with \(t_{RoI} = 0.2\). Since SSD boxes are trained to detect objects on pixel-level, they are often too tight to capture the corresponding Stixels (b). By scaling (c) and/or allowing the capture of only partially intersecting Stixels (d), the modified SSD boxes, i.e. \textit{RoI boxes}, can capture most object Stixels}
     \label{fig:filtering}
   \end{figure}

For each bounding box, the Filtering component finds the potential object Stixels in it. In this work, SSD (Single Shot Detection) \cite{liu2016ssd} trained on Cityscapes is utilized as object detector due to its favourable speed compared to other good deep object detectors \cite{huang2017speed}. SSD accounts for occlusions and predicts tight bounding boxes on pixel-level. Hence, as seen in Fig. \ref{fig:filtering}, the rougher filtered Stixels often do not lie completely within the SSD box. To account for these inaccuracies, two options arise: \begin{enumerate*}[label=(\roman*)] \item Scale the SSD box by a factor \(sc_{RoI} \geq 1\). \item Capture Stixels that only partially overlap with the scaled SSD box. If more than \( t_{RoI} \in (0,1]\) of the Stixel area lies within the scaled SSD box, the Stixel gets captured.\end{enumerate*} The modified SSD box is referred to as {\it RoI box}.

\begin{figure*}[b]
\centering
\includegraphics[width=0.95\linewidth]{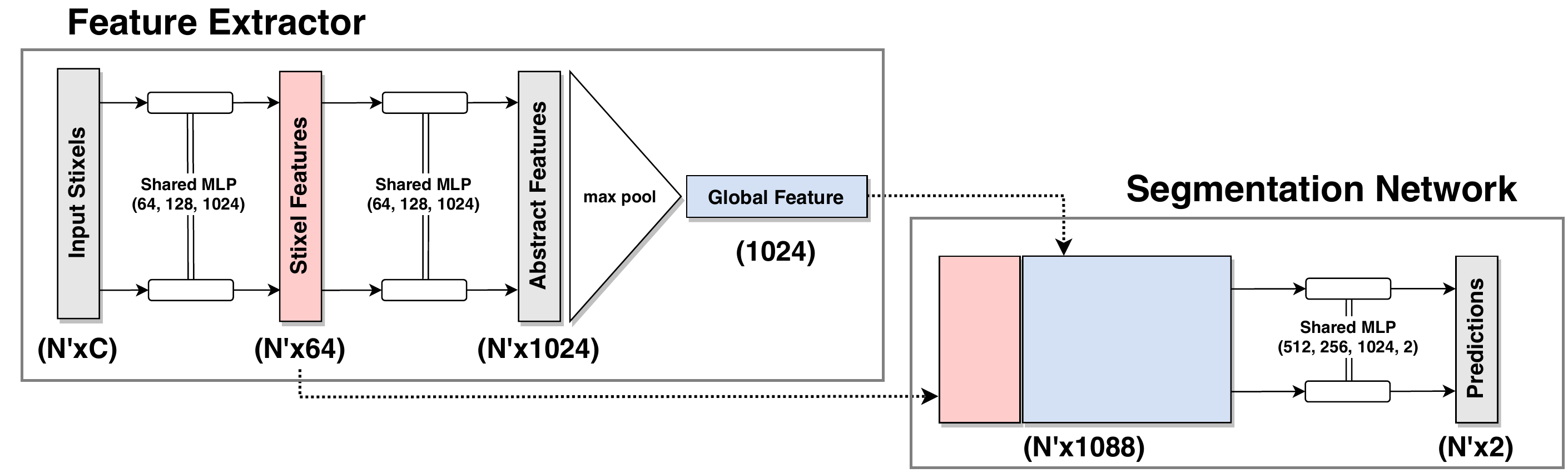}
\caption{The PointNet model\cite{qi2017pointnet} utilized by the StixelPointNet pipeline. The model learns a binary segmentation for the $N'$ filtered Stixels of a RoI box. Shared MLPs extract features for each Stixel. Spatial context is acquired through the global feature, whose shape is invariant to $N'$ due to the use of a symmetric function. The Stixel features and the global feature are concatenated in the Segmentation Network. Here, for each Stixel its binary segmentation is predicted. The probability of a Stixel to belong to the detected instance is called \it{prediction confidence} }
\label{fig:StixelPointNet}
\end{figure*}

\subsection{PointNet Model}
Given the filtered Stixels of a RoI box as input, the model is expected to solve two tasks: \begin{enumerate*}[label=(\roman*)] \item Identify the instance associated to the RoI box. \item Segment the Stixels of the RoI box that make up this instance. \end{enumerate*} 
As noted in the introduction, Stixels can be regarded as points in a point cloud. Hence, the utilized model as shown in Fig. \ref{fig:StixelPointNet} closely resembles a PointNet architecture \cite{qi2017pointnet} with the one exception that its STNs (Spatial Transform Networks) are omitted due to performance gains.

\subsection{Best Prediction Selection}
\label{m:nms}
Considering the Stixel-wise output of the model, one of three cases arises for each Stixel in the input image. \textbf{1)} It has never been seen in any RoI box and therefore does not have any prediction confidence. \textbf{2)} It has been seen in one RoI box and therefore has one prediction confidence. \textbf{3)} It has been seen in multiple RoI boxes and therefore has multiple prediction confidences. The BPS component assigns to each Stixel an instance ID. In case 1), the Stixel is assigned the ID of the background label. In case 2), the Stixel is assigned the available instance ID if its prediction confidence exceeds a confidence threshold \(t_{conf}\). Otherwise it is assigned to the background. In case 3), if multiple Stixels should exceed \(t_{conf}\), the best predictions has to be selected based on each Stixels‘ prediction confidence and RoI box confidence \(c_{bb}\). Neither confidence alone solves the selection problem: The predicition confidence of a perfect model would always be $1$, while the RoI box confidence is ignorant of specific Stixels. Hence, a weighted sum of both confidences is used to decide the winning instance ID, i.e. it is decided for confident predictions in confident RoI boxes.

\section{Baselines}
\label{baselines}

Instance segmentation directly on Stixels is a comparatively novel task. While Hehn et al. \cite{hehn2019instance} show the performance of their algorithm on pixel-level, no baseline on Stixel-level currently exists. This section introduces three such baselines.

\begin{figure*}[t]
\begin{minipage}{0.005\linewidth}
$\,$
\end{minipage}
\begin{minipage}{0.445\linewidth}
   \includegraphics[width=0.95\linewidth,valign=T]{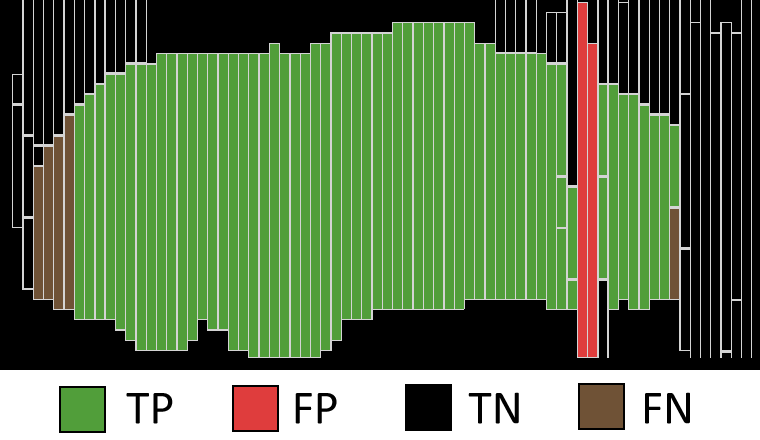}
    \caption{The statistical approach as an alternative to the PointNet model: depicted are the 112 Stixels of one RoI box capturing an instance of class \textit{car}. Since RoI boxes for cars on average contain 57 \% object Stixels (see Table \ref{table:class-wise-percentages}), the 63 centermost Stixels according to the $L^{2}$ distance get labeled as object Stixels}
     \label{fig:statistical}
\end{minipage}
\begin{minipage}{0.045\linewidth}
    $\,$
\end{minipage}
\begin{minipage}{0.48\linewidth}
     \subfloat[Car clusters\label{fig:HACimg0}]{%
       \includegraphics[width=0.48\linewidth,valign=T]{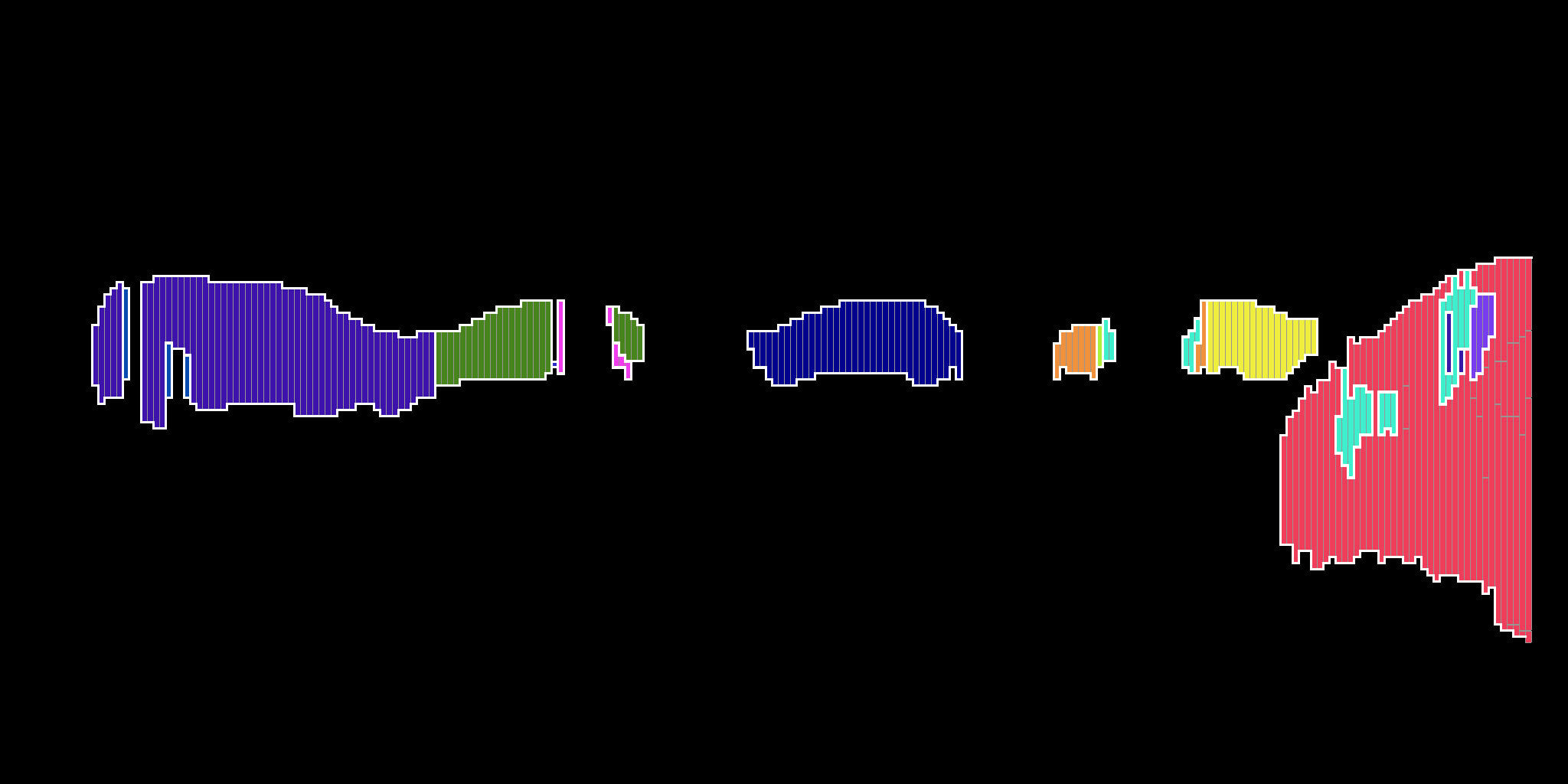}
     }
     \hfill
     \subfloat[Person clusters\label{fig:HACimg1}]{%
        \includegraphics[width=0.48\linewidth,valign=T]{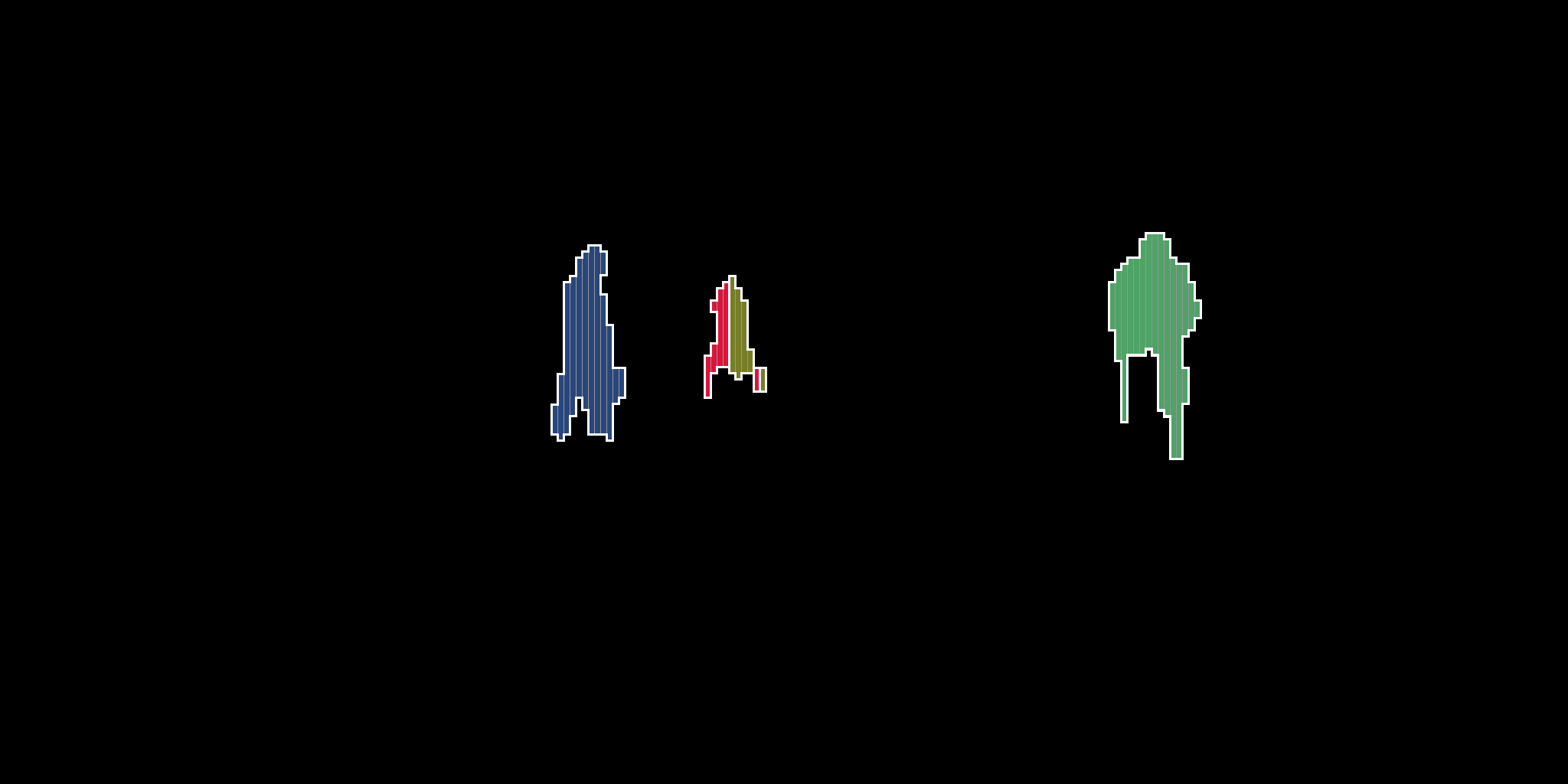}
     }
     \\
     \subfloat[Bike cluster\label{fig:HACimg2}]{%
       \includegraphics[width=0.48\linewidth]{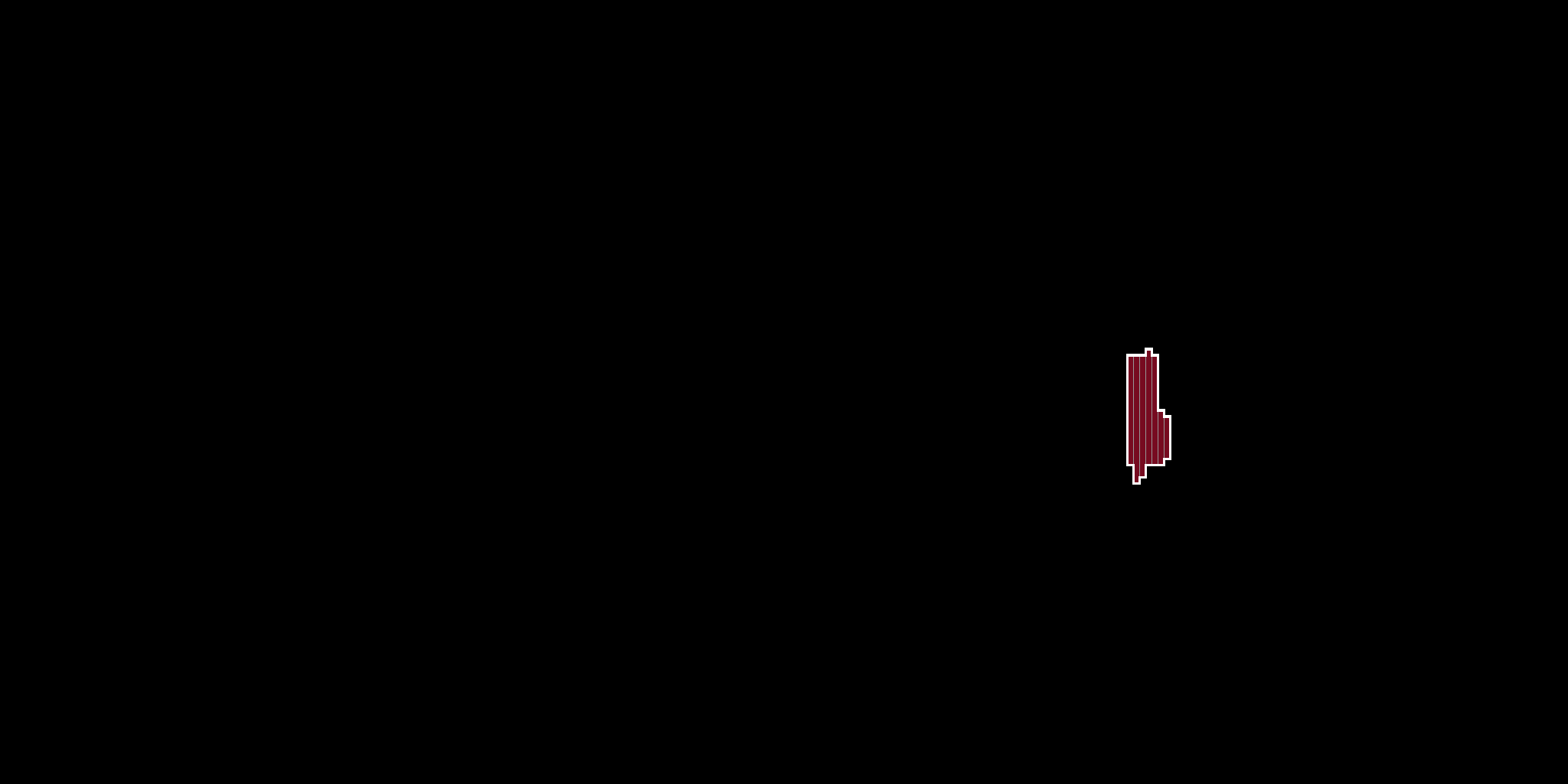}
     }
     \hfill
     \subfloat[Segmentation\label{fig:HACimg3}]{%
        \includegraphics[width=0.48\linewidth]{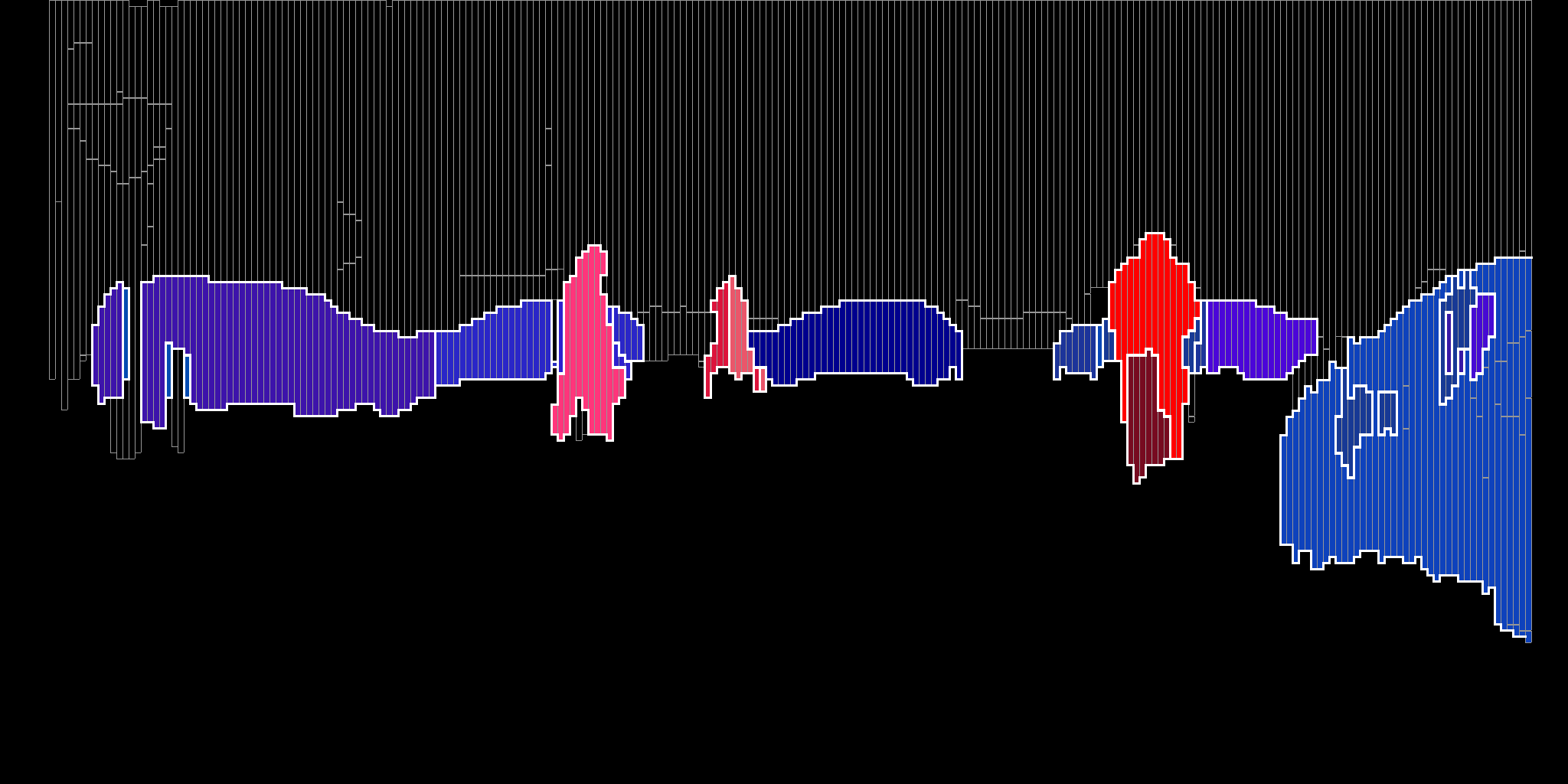}
     }
   \caption{HAC$_{img}$ as an alternative to StixelPointNet. HAC$_{img}$ clusters the Stixels of each class separately. For the final segmentation, each cluster (illustrated by Stixels with the same color) constitutes a predicted instance}
\label{fig:HACimg}
\end{minipage}
\end{figure*}

\setlength{\tabcolsep}{4pt}
\begin{table}[b]
\begin{center}
\caption{ Class-wise instance percentages \(p_{c}\) for the filtering parameters \(sc_{RoI} = 1.0 \) and \(t_{RoI} = 0.1\) (compare Sec. \ref{m:filtering}) on the Stixel-level dataset \(SD_{0.35}\) (see Sec. \ref{eval:datasets}). When an instance of class \(c\) is associated to the RoI box, its captured object Stixels on average make up \(p_{c}\) of the RoI box Stixels}
\label{table:class-wise-percentages}
\begin{tabular}{l |c c c c c c c}
& pers. & rider & car & truck & bus & m.bike & bike\\
\hline
\(p_{c}\) for Train Dataset & .52 & .45 & .58 & .56 & .56 & .51 & .46 \\
\(p_{c}\) for Validation Dataset & .51 & .43 & .57 & .57 & .59 & .54 & .44 \\
\end{tabular}
\end{center}
\end{table}
\setlength{\tabcolsep}{1.4pt}

\subsection{Statistical Approach}

In each RoI box, a certain percentage $p$ of Stixels belong to the detected object. Analyzing $p$ for each class in a given Stixel-level dataset yields the class-wise average percentage of object Stixels $p_{c}$ (see Table \ref{table:class-wise-percentages}). Using this knowledge, the PointNet model can be replaced by a simple statistical approach that labels $p_{c}$ Stixels as object Stixels in a RoI box of class $c$. As it is reasonable to assume that the detected object is situated near the center of the RoI box, Stixels are ordered either by their $L^{1}$ or $L^{2}$ distance to the RoI box center. Those Stixels closest to the RoI box center get labeled first as depicted in Fig. \ref{fig:statistical}.

\subsection{Hierarchical Clustering}
HAC (Hierarchical Agglomerative Clustering) has long been a tool in segmenting data points as it requires little a-priori knowledge of the data \cite{murtagh1983survey}. In 2015, Dohan et al. \cite{dohan2015learning} used a combination of an object classifier and HAC to perform fast semantic segmentation on Google Street View LIDAR point clouds. For this work, two HAC approaches are considered. HAC$_{RoI}$ serves as an alternative to the PointNet model and as such yields a binary segmentation for the Stixels in each RoI box. HAC$_{img}$ serves as an alternative to the whole StixelPointNet pipeline and as such yields an instance segmented Stixel-image. In general, Stixels are clustered based on their semantic class $l$ as well as on a subset of their 3D coordinates {\it x, y, z}.

For HAC$_{img}$, the input is the Stixel image. In this, each semantic class is clustered separately as depicted in Fig \ref{fig:HACimg}. Since the final number of clusters is unknown, a distance threshold $\mu_{c}$ for each class $c$ is utilized to stop the clustering. Each cluster is then considered a predicted Stixel-instance.
For HAC$_{RoI}$, the input, identical to the PointNet model, is the Stixels of one RoI box. Only Stixels with the same semantic label $l$ as the predicted RoI box class $l_{bb}$ are clustered. For this, the same $\mu_{c}$ as for HAC$_{img}$ is used. Since each RoI box captures no more than one instance, only the cluster with the maximum area is chosen as the winning binary segmentation.

   \begin{figure*}[t]
     \subfloat[Stixels projected onto instance gt\label{fig:gt_gen1}]{%
       \includegraphics[width=0.48\linewidth]{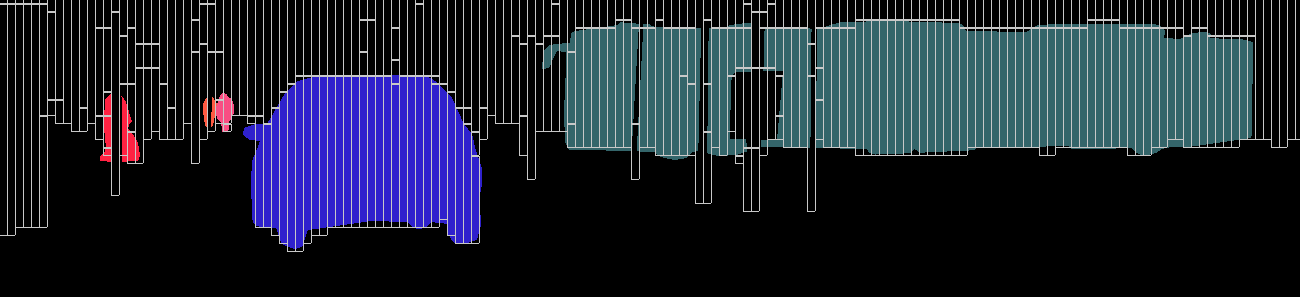}
     }
     \hfill
     \subfloat[The resulting Stixel-level instance gt\label{fig:gt_gen2}]{%
        \includegraphics[width=0.48\linewidth]{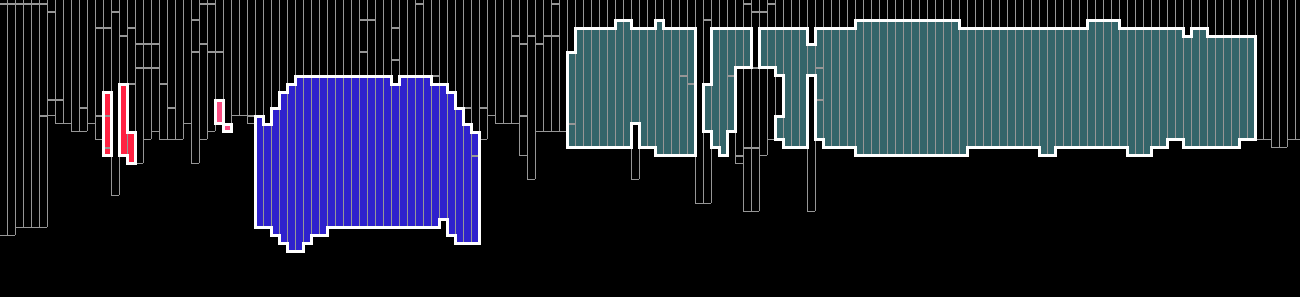}
     }
     \caption{Generating Stixel-level instance ground truth from pixel-level instance ground truth. By projecting Stixels onto the corresponding Cityscape's pixel-level instance ground truth image, a Stixel {\it Stx} is assigned to an instance {\it I} if its overlap \(\frac{area_{Stx}  \cap  area_{I}}{area_{Stx}}\) with that instance exceeds an assignment threshold \(t_{ov}\). If the overlap exceeds \(t_{ov}\) for multiple instances, the instance with the maximum overlap is chosen. If the overlap does not exceed \(t_{ov}\) for any instance, it is considered background. This can be observed for the far away middle pedestrian (red), who is not accurately represented by any Stixels}
     \label{fig:gt_gen}
   \end{figure*}

\section{Evaluation}

\subsection{Datasets and Metrics}
\label{eval:datasets}
Evaluations were performed on the Cityscapes benchmark \cite{cordts2016cityscapes} (1525 images) and two derived datasets: the Cityscapes validation dataset (500 images) and \(SD_{0.35}\) (495 images), a Stixel-level dataset generated from Cityscapes’ validation dataset.

\(SD_{0.35}\) is generated as depicted in Fig. \ref{fig:gt_gen}. To assess how well the generated Stixel-level ground truth represents Cityscapes' pixel-level annotations, the former can be considered a prediction for the latter, i.e. an AP (Average Precision) metric can be calculated. Doing this showed that the maximum AP for all classes is achieved for an overlap threshold around \(t_{ov} = 0.35\). Hence, this is the chosen threshold for \(SD_{0.35}\). All three datasets contain instances of 10 different object classes, 8 of which are used for evaluation.

\begin{figure}[t]
     \subfloat[\label{fig:SDopt_gt}]{%
       \includegraphics[width=0.3\linewidth]{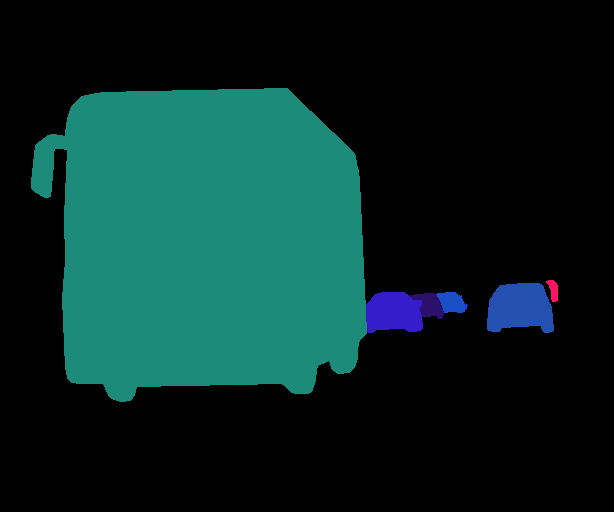}
     }
     \hfill
     \subfloat[\label{fig:SDopt_035}]{%
        \includegraphics[width=0.3\linewidth]{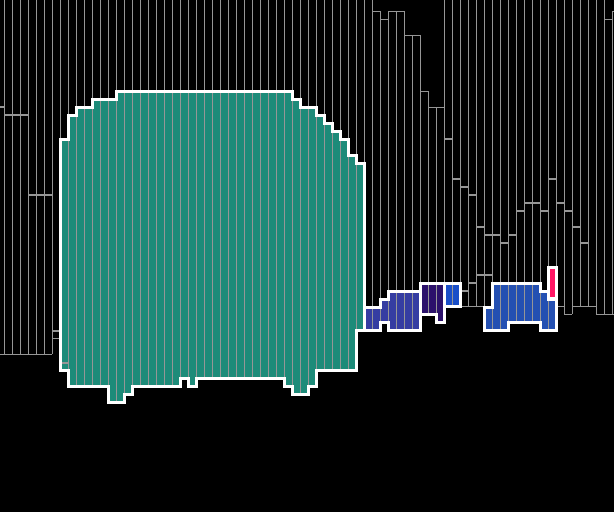}
     }
     \hfill
     \subfloat[\label{fig:SDopt_opt}]{%
        \includegraphics[width=0.3\linewidth]{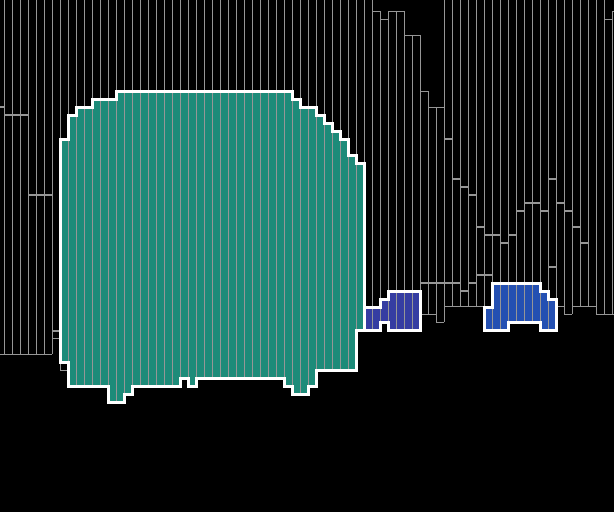}
     }
     \caption{From Cityscapes' pixel-level annotations (a) the Stixel-level ground truth \(SD_{0.35}\) is generated (b). (c) shows the pixel-level TP matches, when using (b) to predict (a). Here, only 3 of the 6 Stixel-level ground truth instances can be matched to their corresponding pixel-level instance. For the remaining instances, the Stixel representation is too rough to achieve an $IoU > 0.5$ as required for positive matching. Hence, any algorithm that successfully predicts those 3 instances on Stixel-level will be penalized with additional false positives in its pixel-level AP results. This explains why Stixel-level predictions lead to low pixel-level AP scores}
     \label{fig:SDopt}
   \end{figure}

The availability of Stixel-level annotations not only for training, but also for evaluation is vital. Evaluating Stixel-level segmentations on pixel-level annotations is a suboptimal procedure due to the simplified Stixel shape yielding less accurate results by default. This goes so far as correct Stixel-level predictions for very small or highly occluded instances actually decreasing pixel-level AP, since the Stixel representation does not reach the required pixel-level IoU to map prediction and pixel annotation. As illustrated and explained in Fig. \ref{fig:SDopt}, this applies even to instances which on Stixel-level are still easily identifiable to the human eye.

Still, for comparison with Hehn et al.\cite{hehn2019instance} as well as for providing context on the Cityscapes benchmark we also present pixel-level AP scores.

\setlength{\tabcolsep}{2.85pt}
\begin{table}[b]
	\begin{center}
    \caption{Results on the \textbf{Stixel-level} \(SD_{0.35}\) dataset based on Cityscapes' validation dataset. None of the approaches score any AP for the class \textit{train}, since neither the semantic class of the Stixels nor the semantic class of the used SSD boxes differentiates between busses and trains. Still, AP and AP$_{50\%}$ reflect the mean of all 8 classes}
    \label{table:SD035ResultsClasses}
    \begin{tabular}{l | c |c | c c c c c c c}
    & AP & AP$_{50\%}$ & \multicolumn{7}{|c}{class-wise AP$_{50\%}$} \\
    method & & & pers. & rider & car & truck & bus & m.bike & bike\\
    \hline
    Statistical & 15.1 & 35.2 & 47.8 & 30.1 & 57.4 & \textbf{36.8} & 55.0  & 26.6 & 28.3\\
    HAC$_{RoI}$ & 21.2 & 37.7 & 43.0 & 42.2 & 47.9 & 34.1 & 52.1  & \textbf{38.8} & \textbf{43.2}\\ 
    HAC$_{img}$ & 19.4 & 31.7 & 40.4 & 31.4 & 45.2 & 25.1 & 41.8  & 31.9 & 37.7\\
    StxPN (our)& \textbf{25.4} & \textbf{39.6} & \textbf{49.5} & \textbf{45.4} & \textbf{61.2} & 33.6 & \textbf{55.7}  & 27.8 & \textbf{43.2}\\
    \end{tabular}
    \end{center}
\end{table}
\setlength{\tabcolsep}{1.4pt}

\subsection{Feature Selection}
While the input Stixel image consists of Semantic Stixels with features as laid out in Section \ref{m:}, for the instance segmentation task the following modifications are applied to these features: during the filtering stage the absolute 2D features $u_{tl},\: v_{tl},\: u_{br},\: v_{br}$ are substituted by the normalized position and height {\it u', v', h'} inside the RoI box. In addition, the label confidence \(c\) is neglected in favor of the bounding box label \(l_{bb}\) to cope with highly occluded objects, resulting in a feature vector $Stx' = [x,\: y,\: z,\: w,\: h,\: u',\: v',\: h',\: l,\: l_{bb}]$. 

\subsection{Hyperparameters}
For the Statistical approach, the $L^{2}$ distance performed slightly better than the $L^{1}$ distance, so $L^{2}$ was chosen. For both HAC algorithms, the complete-linkage criterion in conjunction with the Euclidian distance of only the Stixel's $x$ and $z$ position performed best. The chosen distance thresholds in meters are ${\mu_{person}=1}$, ${\mu_{rider}= 5}$, ${\mu_{car}= 5}$, ${\mu_{truck}= 15}$, ${\mu_{bus}= 20}$, ${\mu_{m.bike}= 5}$ and ${\mu_{bike}= 2.5}$. For HAC$_{RoI}$, the cluster with the maximum Stixel area is chosen as the winning cluster, as it performed slightly better than choosing the cluster with the maximum number of Stixels. For StixelPointNet, the chosen parameters for each component are given below.
\begin{description}
    \item[Filtering:] The filtering component uses an overlap threshold of $t_{RoI} = 0.1$ and a RoI box scale of $sc_{RoI} = 1$.
    \item[Model:] The model is a PointNet without STNs. The MLP layer sizes can be inferred from Fig. \ref{fig:StixelPointNet}. The model loss is cross entropy. A batch size of 32 and the ADAM optimizer with an initial learning rate of 3e-3 are utilized.
    \item[BPS:] For BPS we use a confidence threshold of $t_{conf} = 0.5$ and decide the winning id via the max of \( 0.75 \cdot c_{bb} + 0.25 \cdot pc \), where $pc$ is prediction confidence.
    \end{description}

\subsection{Results}
\label{results}

\begin{figure*}[t]
     \subfloat[Ground Truth\label{fig:results0}]{%
       \includegraphics[width=0.32\linewidth]{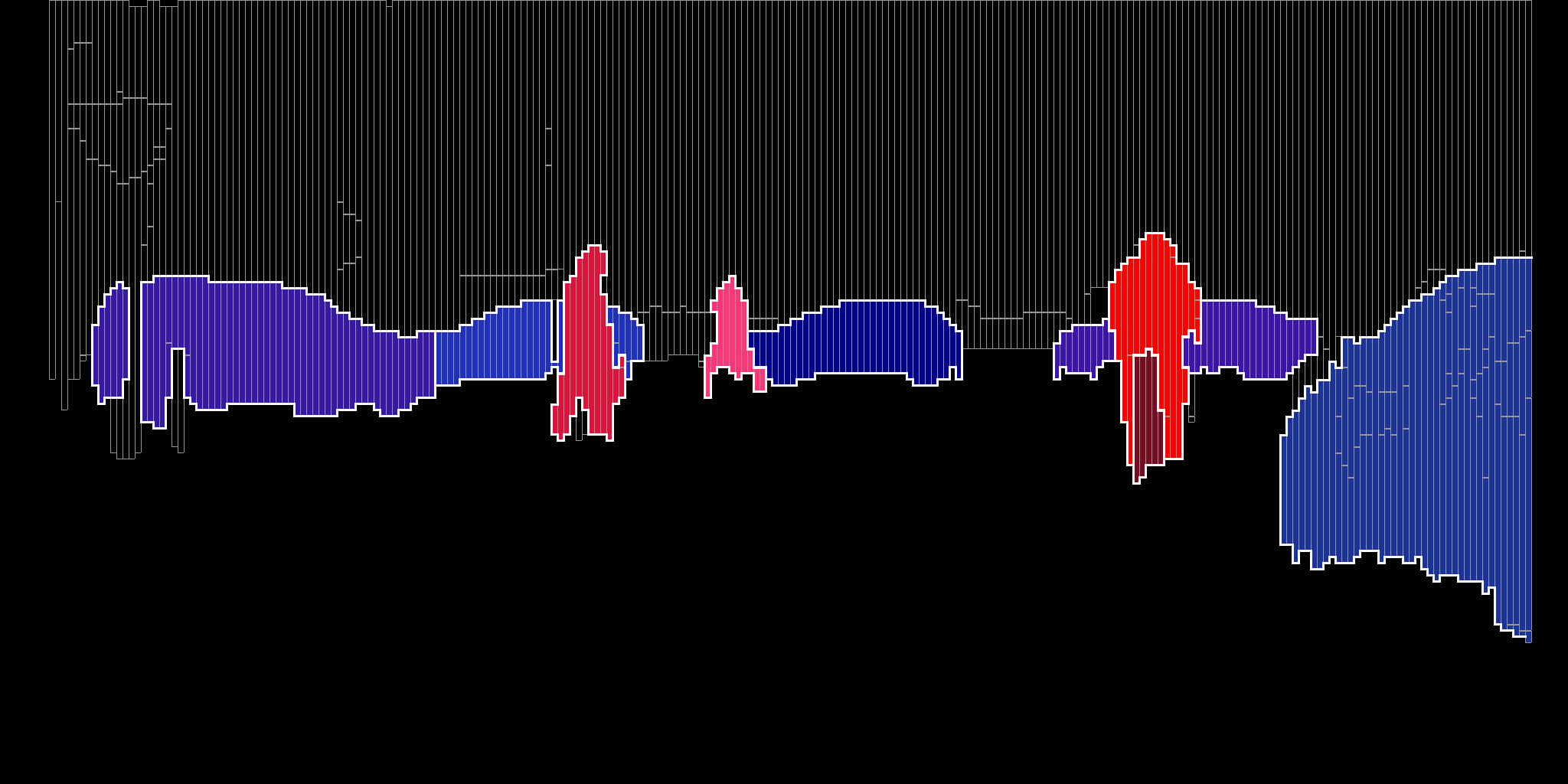}
     }
     \hfill
     \subfloat[Statistical $L^{1}$\label{fig:results1}]{%
        \includegraphics[width=0.32\linewidth]{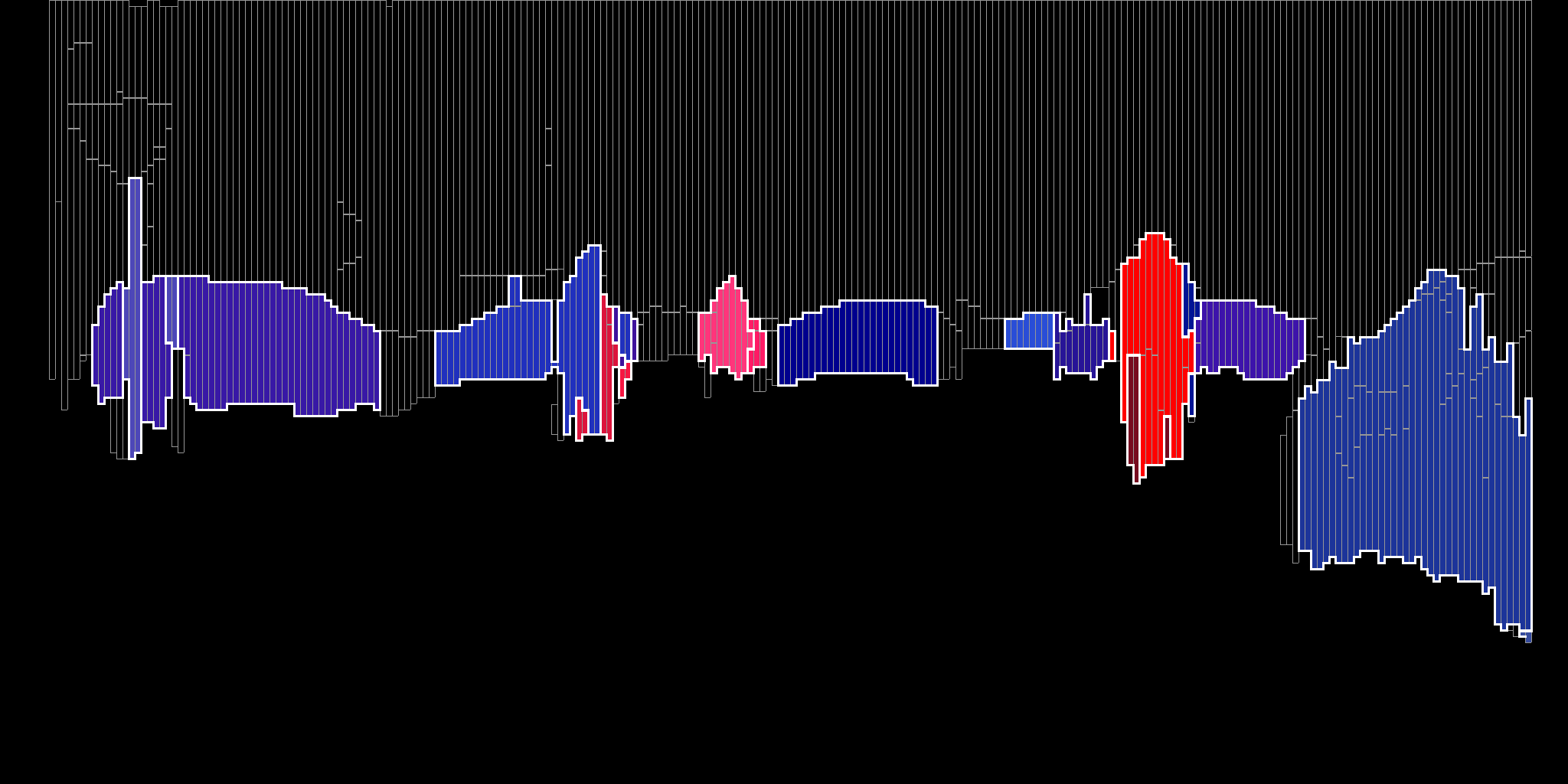}
     }
     \hfill
     \subfloat[Statistical $L^{2}$\label{fig:results2}]{%
        \includegraphics[width=0.32\linewidth]{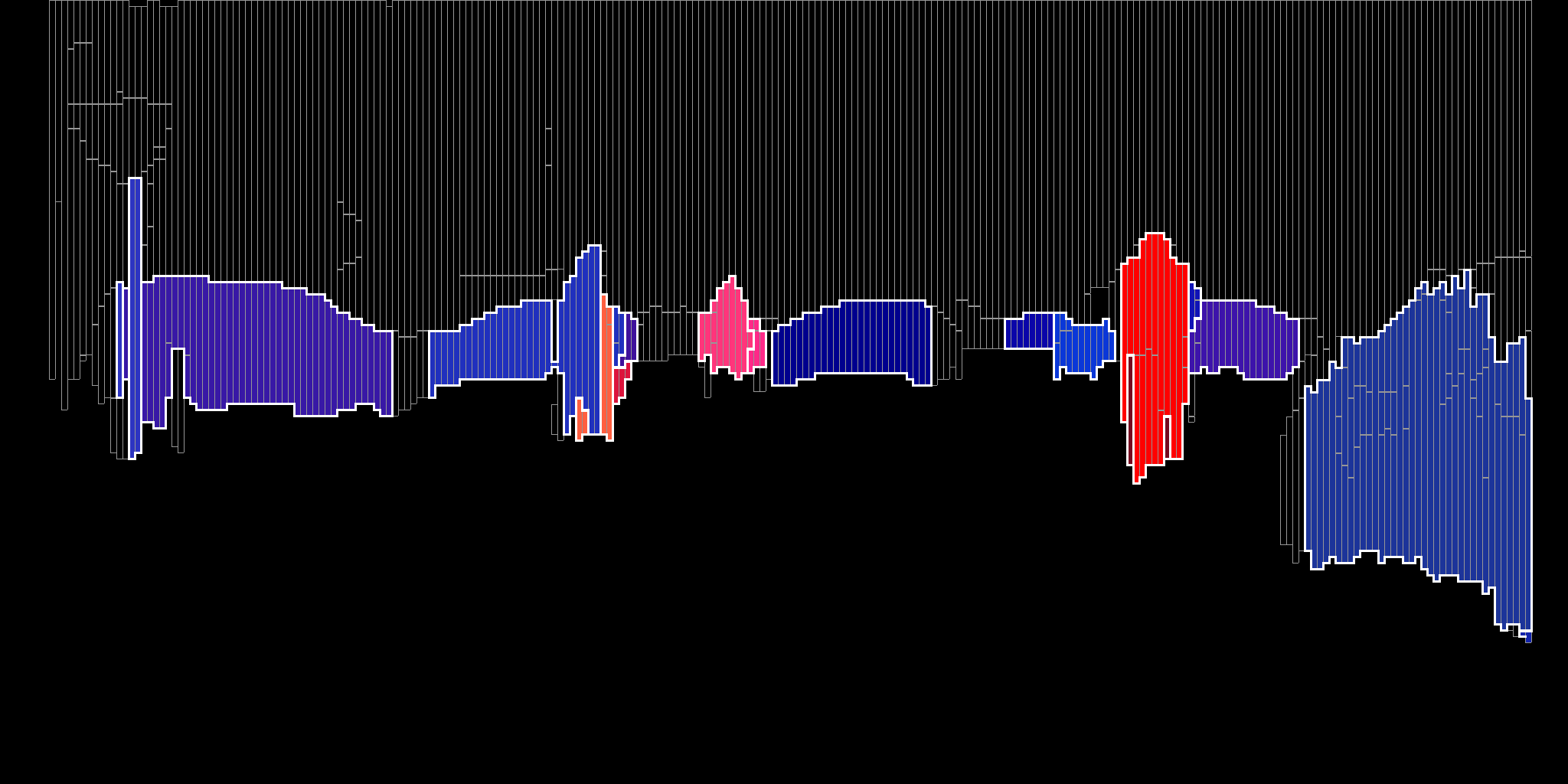}
     }\\
     \subfloat[StixelPointNet\label{fig:results3}]{%
       \includegraphics[width=0.32\linewidth]{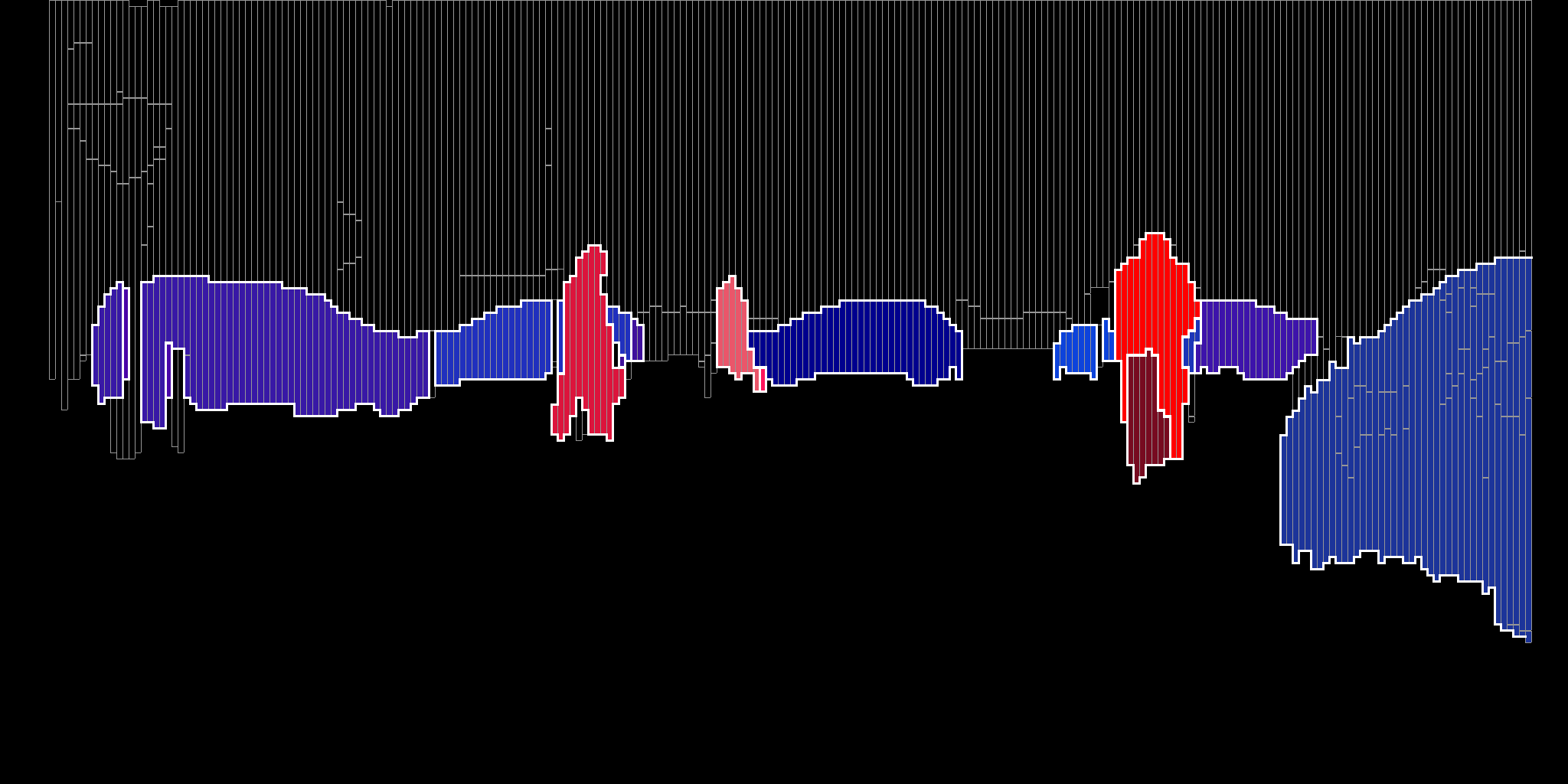}
     }
     \hfill
     \subfloat[HAC$_{RoI}$\label{fig:results4}]{%
        \includegraphics[width=0.32\linewidth]{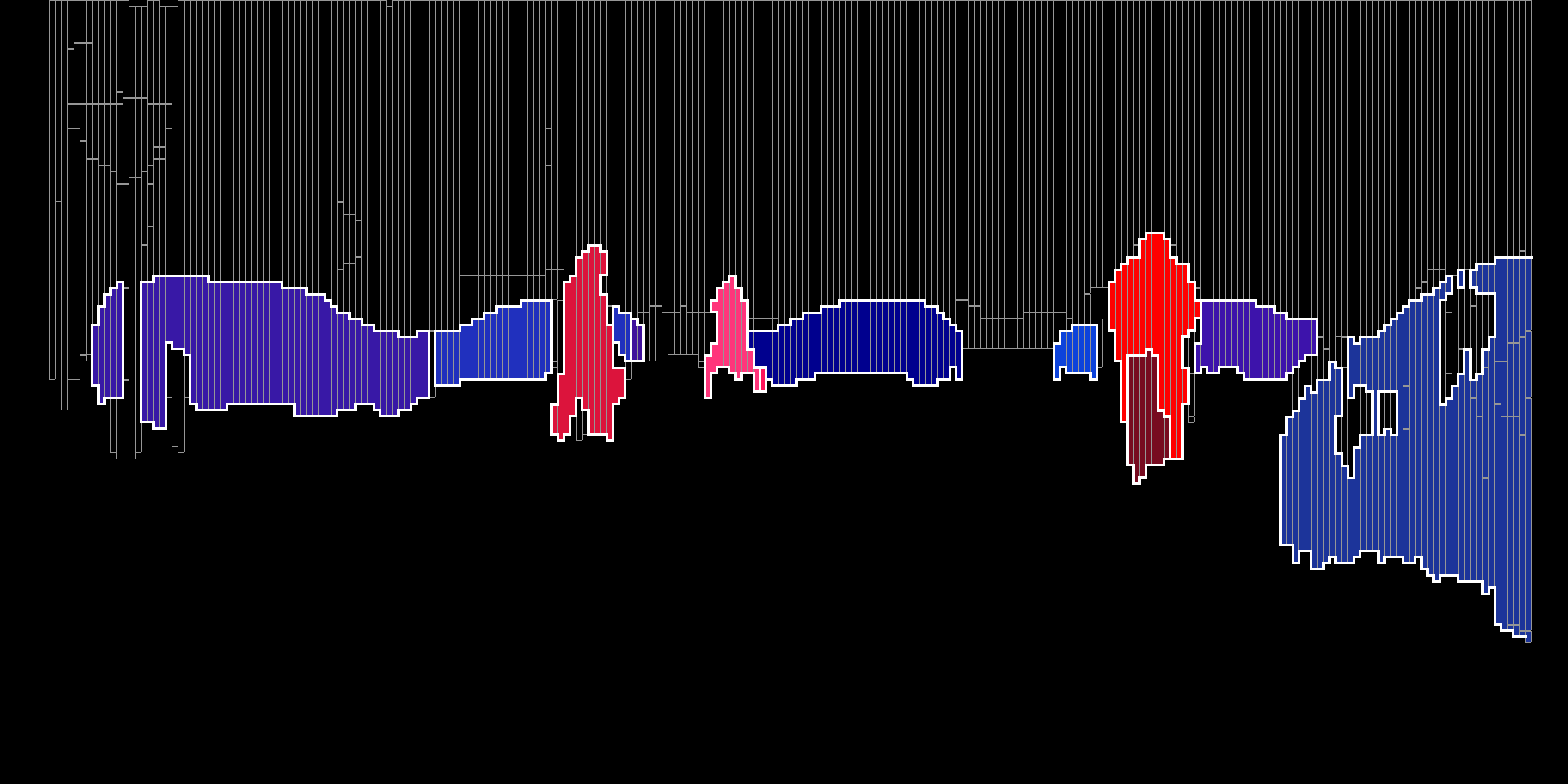}
     }
     \hfill
     \subfloat[HAC$_{img}$\label{fig:results5}]{%
        \includegraphics[width=0.32\linewidth]{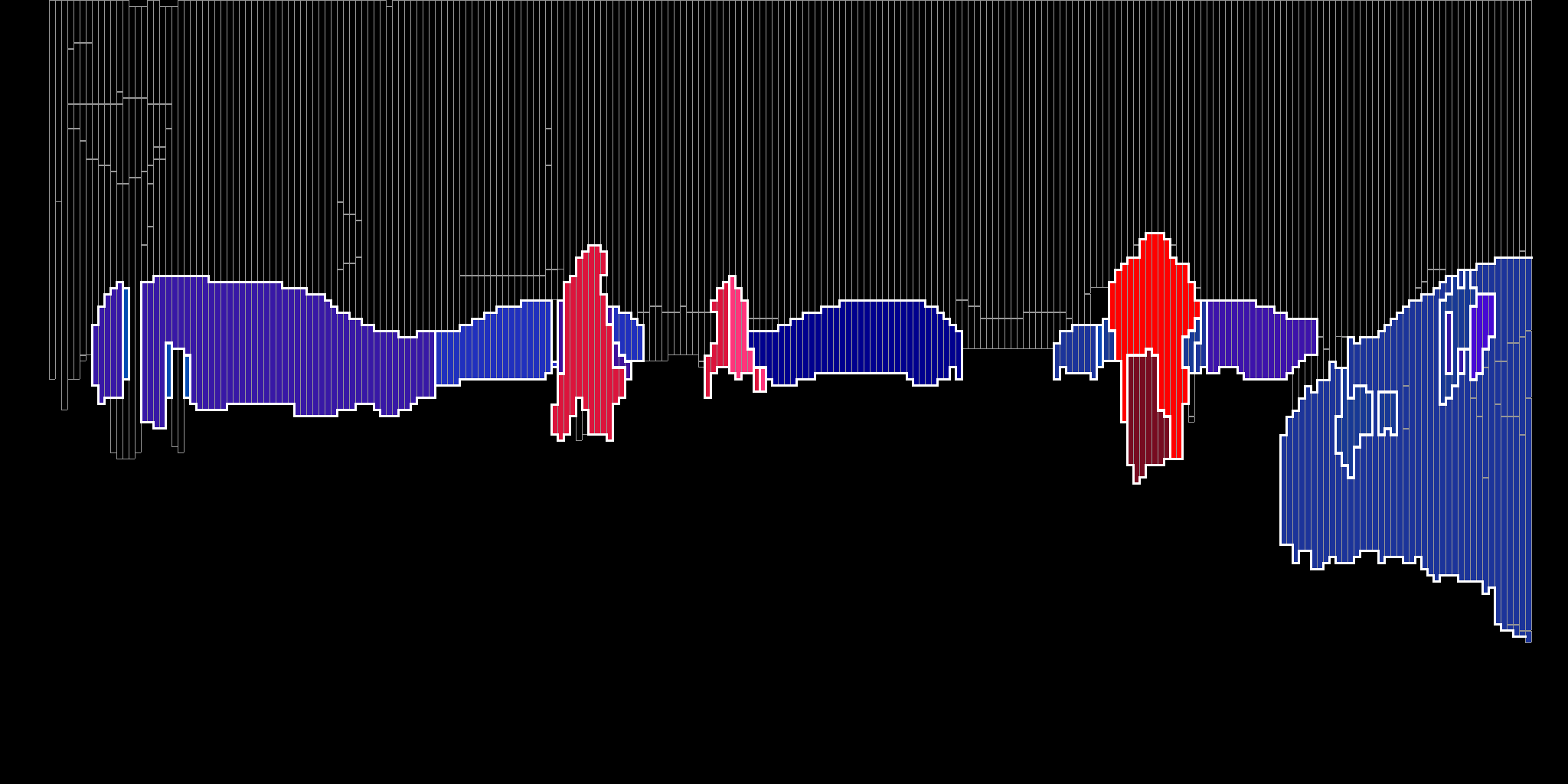}
     }
     \caption{Instance segmentation of all proposed methods. StixelPointNet achieves the most accurate results, but like all other methods splits the right car behind the cyclist. The statistical approach, regardless of the chosen distance metric, struggles with occlusions and big instances, since the latter often contain a higher percentage of object Stixels. HAC$_{RoI}$ produces some empty artefacts for the rightmost car. For HAC$_{img}$, those same artefacts are still present, but are now a predicted instance, resulting in additional FP predictions
     }
     \label{fig:results}
   \end{figure*}

\begin{figure}[t]
    \centering
     \subfloat[\(SD_{0.35}\) ground truth \label{Val2_0}]{%
     \includegraphics[width=0.80\linewidth]{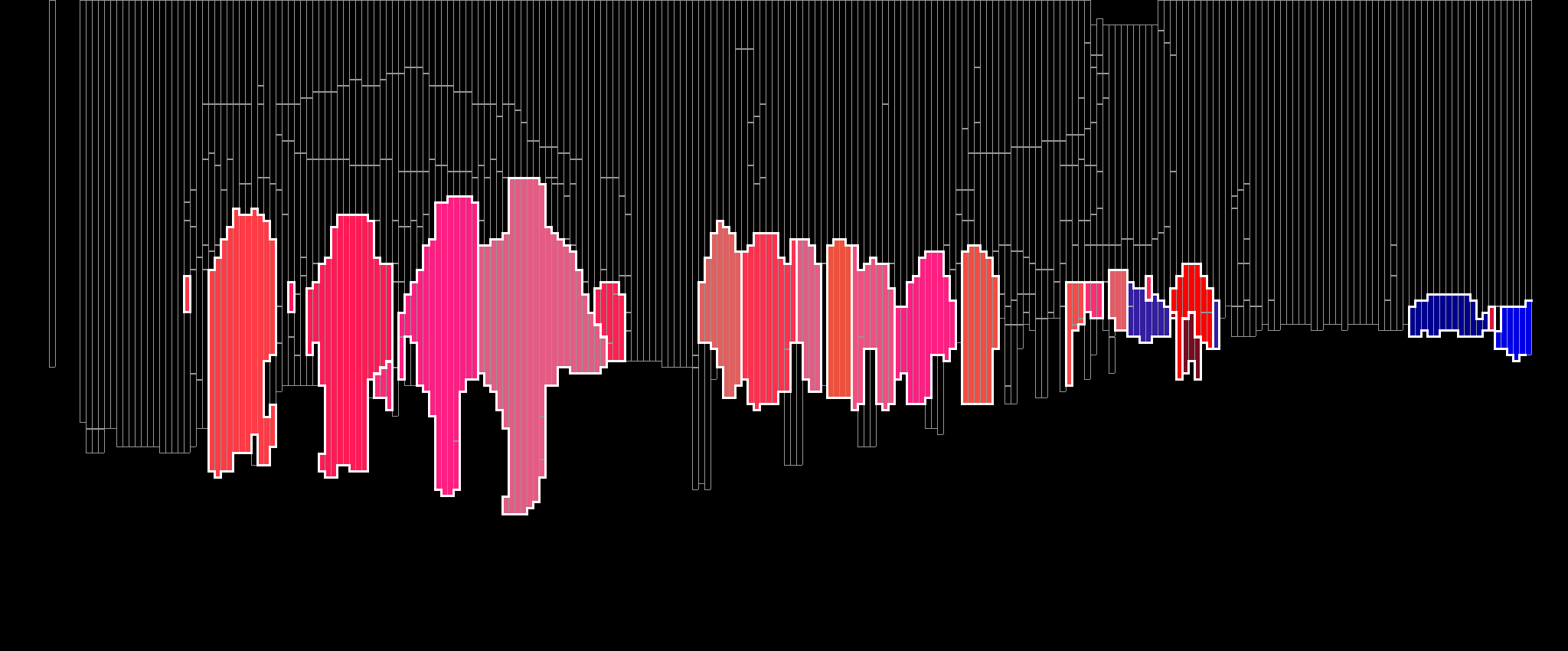}
     }
     \\
     \subfloat[StixelPointNet segmentation \label{Val2_1}]{%
     \includegraphics[width=0.80\linewidth]{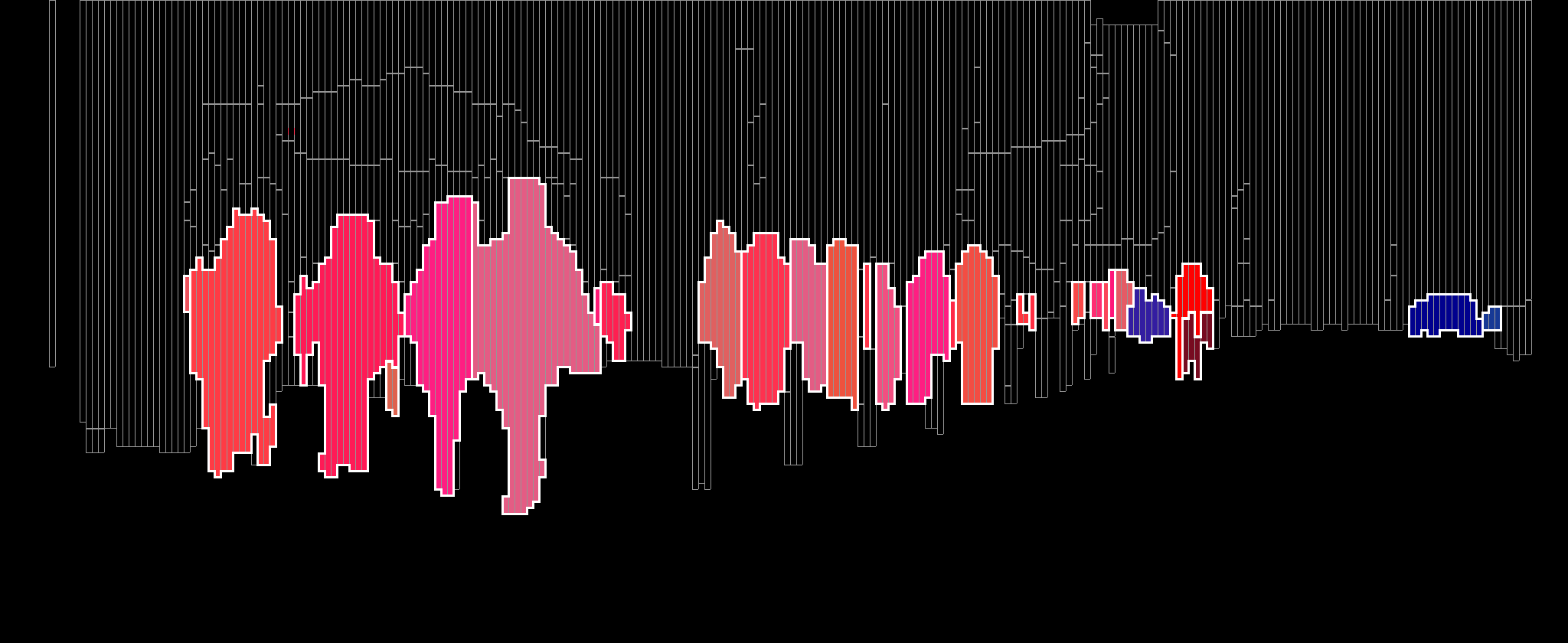}
     }
     \caption{A difficult scene from the Cityscapes validation dataset with many pedestrians}
     \label{Val2}
\end{figure}

\begin{figure}[t]
    \centering
     \subfloat[RGB image \label{Test0_0}]{%
     \includegraphics[width=0.80\linewidth]{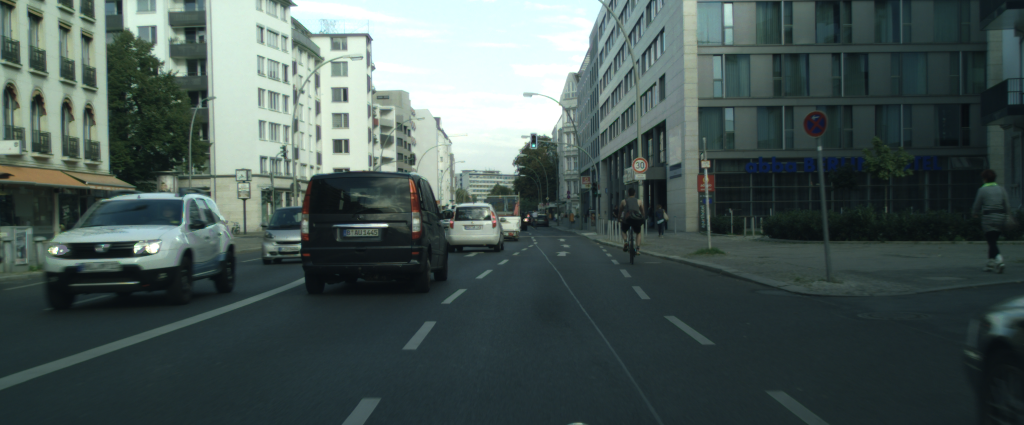}
     }
     \\
     \subfloat[StixelPointNet segmentation \label{Test0_1}]{%
     \includegraphics[width=0.80\linewidth]{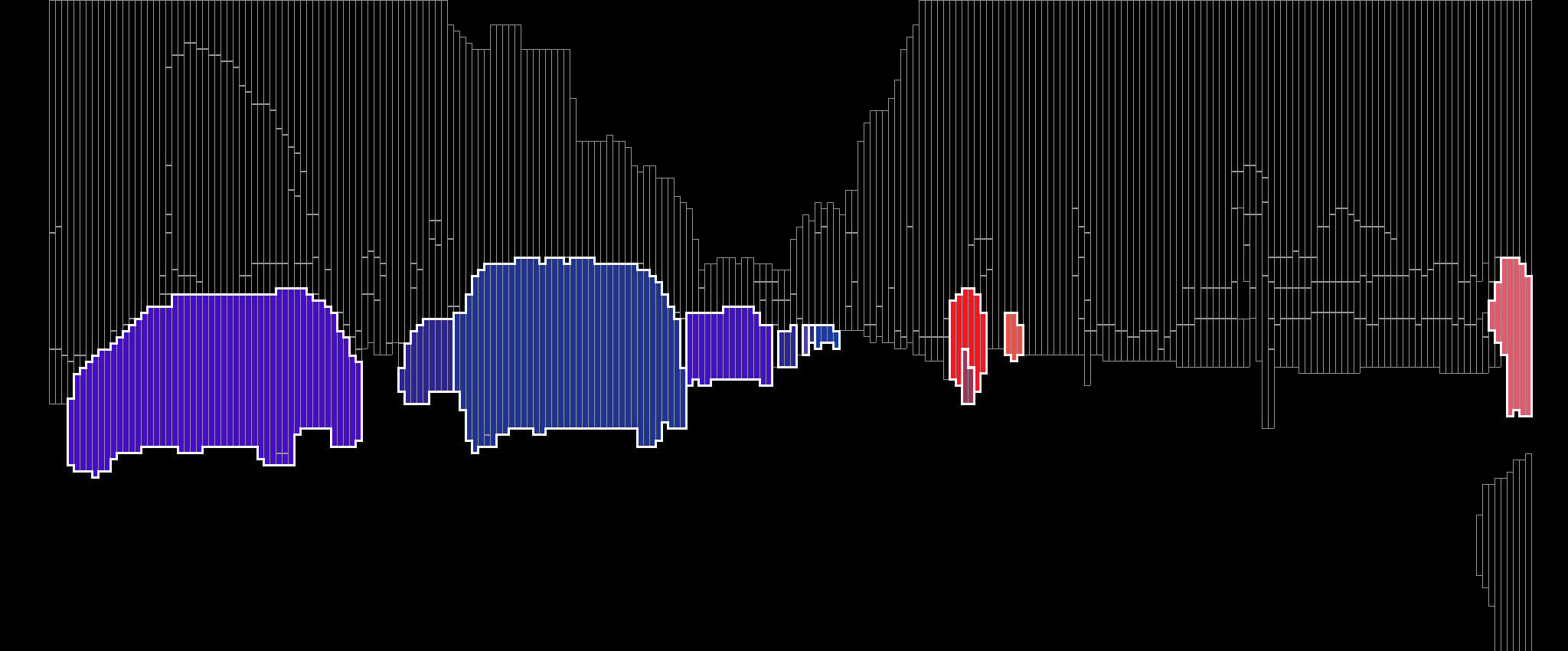}
     }
     \caption{A scene from the Cityscapes test dataset}
     \label{Test0}
\end{figure}

Table \ref{table:SD035ResultsClasses} depicts the performance of StixelPointNet and all proposed baselines on the \(SD_{0.35}\) Stixel-level dataset. Please note that none of the proposed methods can predict the class \textit{train} as the SSD algorithm used for region proposals as well as Semantic Stixels combine the classes \textit{bus} and \textit{train} to one class \textit{large vehicle} in order to compensate for the very limited training examples of class \textit{train}. Hence, their AP score for \textit{train} is 0. Still, StixelPointNet scores the highest AP for 5 of the 7 recognised classes, losing only for the classes \textit{truck} and \textit{motorcycle}. Noticeably, these are classes with very few associated SSD boxes, together making up less than 1.6\% of all training samples. In contrast, for class \textit{car} which makes up ~32\% of the training samples, an AP of over 62\% is scored. A qualitative analysis for a typical scene is given in Fig. \ref{fig:results}. Here, StixelPointNet achieves the cleanest segmentation of all proposed methods. Further qualitative results on the Cityscapes validation and test dataset are given in Fig. \ref{Val2} and \ref{Test0} respectively.

\setlength{\tabcolsep}{12.8pt}
\begin{table*}[t]
\rule{0pt}{2ex} 
\begin{threeparttable}
    \renewcommand{\TPTminimum}{\linewidth} 
    \caption{Results on Cityscapes' \textbf{pixel-level} validation and test dataset. Results for the latter are marked by $^1$. For Instance Stixels \cite{hehn2019instance} no class-wise AP$_{50\%}$ is reported. \textit{nAP}$_{50\%}$ depicts a normalized score, i.e. the actual AP$_{50\%}$ divided by the maximum achievable AP$_{50\%}$ for Stixels as illustrated in Table \ref{table:maxAchievable}. For \cite{hehn2019instance} this is an artificial value since they use different Stixels. 
    }
    \label{table:valResultsClasses}
    \begin{tabular}{l | c |c | c | c c c c c c c}

    & AP & AP$_{50\%}$ & \textit{nAP}$_{50\%}$ & \multicolumn{7}{|c}{class-wise AP$_{50\%}$} \\
    method & & & & pers. & rider & car & truck & bus & m.bike & bike\\
    \hline
    Statistical & 6.6 & 19.6 & .38 & 22.3 & 11.1 & 32.0 & \textbf{31.1} & 47.3 & 8.3 & 4.6\\
    HAC$_{RoI}$ & 9.1 & 21.7 & .42 & 19.9 & 25.3 & 26.9 & 23.4 & 46.8 & \textbf{15.1} & \textbf{15.5}\\
    HAC$_{img}$ & 6.3 & 14.2 & .28 & 13.8 & 15.5 & 18.5 & 11.6 & 31.2 & 8.9 & 13.7\\
    IS \cite{hehn2019instance} & 11.4 & 21.8 & (.42) & - & - & - & - & - & - & -\\
    StxPN (our)& \textbf{11.7} & \textbf{24.9} & \textbf{.48} & \textbf{27.2} & \textbf{25.7} & \textbf{39.6} & 27.9 & \textbf{51.3} & 8.9 & 15.1\\
    \hline
    \noalign{\smallskip}
    StxPN (our)$^1$ & 8.5 & 19.3 & - & 24.6 & 24.9 & 31.8 & 22.3 & 27.3 & 11.6 & 11.4\\
    \end{tabular}
    \begin{tablenotes}\fontsize{5.8}{4}
     \item[1] Submission results: \url{https://www.cityscapes-dataset.com/anonymous-results/?id=f02219876e0960b289a16a5cdf6af477fe81760e158219e608a378bd8aa1c306}
    \end{tablenotes}
\end{threeparttable}
\end{table*}
\setlength{\tabcolsep}{1.4pt}

\setlength{\tabcolsep}{11pt}
\begin{table*}[t]
	\begin{center}
    \caption{Upper bounds of StixelPointNet's performance on Cityscapes' \textbf{pixel-level} validation dataset. \textit{nAP}$_{50\%}$ depicts actual AP$_{50\%}$ divided by the best achievable AP$_{50\%}$ for Stixels. \textit{Oracle} describes the AP if no trained model is used, but the binary segmentation ground truth for each RoI box is passed through instead. \textit{Upper Bound} describes the AP achieved when using the \(SD_{0.35}\) Stixel-level ground truth dataset as prediction. Note that for \textit{car} the oracle achieves higher AP than Stixel ground truth due to SSD missing small Stixel-instances that will always result in FP pixel-level predictions
    }
    \label{table:maxAchievable}
    \begin{tabular}{l | c |c | c | c c c c c c c c}
    & AP & AP$_{50\%}$ & \textit{nAP}$_{50\%}$ & \multicolumn{8}{|c}{class-wise AP$_{50\%}$} \\
    method & & & & pers. & rider & car & truck & bus & train & m.bike & bike\\
    \hline
    StxPN (our)& 11.7 & 24.9 & .48 & 27.2 & 25.7 & 39.6 & 27.9 & 51.3 & 0 & 8.9 & 15.1\\
    \hline
    Oracle & 17.8 & 39.2 & .76 & 42.2 & 43.6 & 52.9 & 45.1 & 65.6 & 0 & 29.9 & 34.4\\
    Upper Bound & 23.3 & 51.4 & 1 & 44.5 & 49.0 & 52.7 & 52.2 & 67.2 & 65.5 & 40.9 & 38.9\\
    \end{tabular}
    \end{center}
\end{table*}
\setlength{\tabcolsep}{1.4pt}

Hehn et al. \cite{hehn2019instance} show the performance of their approach (see Sec. \ref{related:onStx}) on Cityscapes' pixel-level validation dataset. Table \ref{table:valResultsClasses} gives the corresponding results of StixelPointNet and all proposed baselines as well as StixelPointNet‘s results on the actual benchmark. Despite Instance Stixels predicting all 8 classes and being able to respect explicit object boundaries during Stixel generation, StixelPointNet achieves 3\% higher AP$_{50\%}$. This indicates that depth as used for Semantic Stixels already contains most information needed to detect object boundaries and shows the strength of learning directly on Stixels.

In Sec. \ref{eval:datasets} we illustrated the inherent problem with comparing Stixel-level predictions on pixel-level annotations. Our concerns are supported by the data of Table \ref{table:maxAchievable}. Here, the failure of SSD to locate certain instances can actually increase achievable pixel-level AP compared to the Stixel-level ground truth. This reaffirms the need for Stixel-level annotations. Table \ref{table:maxAchievable} also shows that StixelPointNet achieves almost $50\%$ of the achievable pixel-level AP$_{50\%}$ score.

\subsection{Runtime Estimation}
\label{runtime}

\setlength{\tabcolsep}{4.8pt}
\begin{table}[b]
	\begin{center}
    \caption{Inference times of StixelPointnet depending on the number of Stixels and SSD boxes. Times are averaged over 1000 runs. \cite{xiong2019upsnet} and \cite{peng2020deep} are pixel-based approaches on cityscapes
    }
    \label{table:runtimes}
    \begin{tabular}{c c c | c | c c}
    Stixels & SSD boxes & features & component & time (ms) & fps\\
    \hline
    753 & 50 & 10 & filtering & 3.7 & 268.3\\
    753 & 50 & 10 & model & 23.1 & 43.4\\
    753 & 50 & 10 & BPS & 1.5 & 664.3 \\
    \hline
    753 & 50 & 10 & overall & 28.4 & 35.2 \\
    1500 & 50 & 10 & overall & 32.2 & 31.1 \\
    753 & 100 & 10 & overall & 32.7 & 30.6 \\
    1500 & 100 & 10 & overall & 35.0 & 28.6 \\
    1500 & 100 & 20 & overall & 35.8 & 27.9 \\
    \hline
    - & - & - & UPS-Net \cite{xiong2019upsnet} & 227 & 4.4 \\
    - & - & - & Deep Snake \cite{peng2020deep} & 217 & 4.7 \\

    \end{tabular}
    \end{center}
\end{table}
\setlength{\tabcolsep}{1.4pt}

We measure inference runtimes on a NVIDIA GeForce GTX 1070 GPU and an Intel Core i7-4790K CPU with 4.00GHz. Analysis of \(SD_{0.35}\) reveals that a Cityscapes image on average is described by 753 Stixels and 50 SSD boxes. To also evaluate runtime under extreme conditions, we synthetically generate Stixels and SSD boxes. We also augment Stixels with an additional number of arbitrary features, to show the influence of that parameter. Note, that to handle all these cases an adjusted, untrained StixelPointNet architecture was used. The according runtimes are given in Table \ref{table:runtimes}.

Table \ref{table:runtimes} shows, that even under extreme conditions StixelPointNet is still performing at 28 fps. Furthermore, as expected, the components filtering and BPS are very fast and hence contribute only little to the overall runtime. 

We compare our average 35.2 fps to the current top 2 pixel-level instance segmentation methods with published runtimes on the Cityscapes benchmark:

\begin{itemize}
    \item UPS-Net \cite{xiong2019upsnet} with 4.4 fps (on NVIDIA GeForce GTX 1080 Ti, Intel Xeon E5-2687W 3.00GHz)
    \item Deep Snake \cite{peng2020deep} with 4.7 fps (on NVIDIA GeForce GTX 1080 Ti, Intel i7 3.7GHz) 
  \end{itemize}


Evidently, StixelPointNet is roughly 7.5 times faster than those pixel-level methods due to working on more compact data.

\section{Conclusion}

This work proposes a novel pipeline called StixelPointNet that uses a PointNet architecture to perform instance segmentation directly on Stixels. In order to provide adequate training data, an algorithm has been developed to generate Stixel-level instance ground truth data from pixel-level annotations. This way, a Stixel-level dataset with over 3300 images for training and validation was created from Cityscapes \cite{cordts2016cityscapes}.

\addtolength{\textheight}{-0.12cm}   

Tests on the Cityscapes validation dataset show that StixelPointNet possesses better instance segmentation capabilities than any of the provided baselines. StixelPointNet also achieves better results than the Instance Stixels method of Hehn et al. \cite{hehn2019instance} which uses pixel-level instance segmentations learned by CNNs to augment Stixels with instance IDs. Our results have been submitted to the Cityscapes instance-segmentation benchmark, constituting a baseline for future Stixel-level approaches. StixelPointNet is also shown to be considerably faster than pixel-based approaches \cite{xiong2019upsnet, peng2020deep} submitted to the same benchmark.

Nevertheless, some areas of improvement present themselves: PointNet has been criticised for having weak generalisation abilities, causing the proposal of several improved architectures \cite{qi2017pointnet++,li2018pointcnn,jiang2018pointsift}, none of which forbid the inclusion of additional features as carried by Stixels. In the future, we plan to implement those more sophisticated architectures into our pipeline to further increase segmentation capabilities, especially in cases where instances are split due to occlusions. Additionally, SSD has been noted to be fast, but better performing object detectors are available \cite{huang2017speed}.  The utilization of a more accurate object detector can reduce the number of false positive predictions for very small, very large or highly occluded instances.

With Deep Learning on unstructured data seeing advancing success, instance segmentation on Stixels specifically and point-based Deep Learning on Stixels generally has become a viable option, opening the door to a compact, yet ever more descriptive perception of the environment as it is required for the future of autonomous driving.

\bibliographystyle{unsrt}
\bibliography{root}

\begin{thebibliography}{10}

\bibitem{badino2009stixel}
Hern{\'a}n Badino, Uwe Franke, and David Pfeiffer.
\newblock The stixel world - a compact medium level representation of the
  3d-world.
\newblock In {\em DAGM}, 2009.

\bibitem{pfeiffer2011towards}
David Pfeiffer and Uwe Franke.
\newblock Towards a global optimal multi-layer stixel representation of dense
  3d data.
\newblock In {\em BMVC}, 2011.

\bibitem{cordts2017stixel}
Marius Cordts, Timo Rehfeld, Lukas Schneider, David Pfeiffer, Markus Enzweiler,
  Stefan Roth, Marc Pollefeys, and Uwe Franke.
\newblock The stixel world: A medium-level representation of traffic scenes.
\newblock {\em Image and Vision Computing}, 68:40--52, 2017.

\bibitem{hirschmuller2005accurate}
Heiko Hirschmuller.
\newblock Accurate and efficient stereo processing by semi-global matching and
  mutual information.
\newblock In {\em CVPR}, 2005.

\bibitem{pfeiffer2010efficient}
David Pfeiffer and Uwe Franke.
\newblock Efficient representation of traffic scenes by means of dynamic
  stixels.
\newblock In {\em IV}, 2010.

\bibitem{pfeiffer2011modeling}
David Pfeiffer and Uwe Franke.
\newblock Modeling dynamic 3d environments by means of the stixel world.
\newblock {\em IEEE Intelligent Transportation Systems Magazine}, 3(3):24--36,
  2011.

\bibitem{schneider2016semantic}
Lukas Schneider, Marius Cordts, Timo Rehfeld, David Pfeiffer, Markus Enzweiler,
  Uwe Franke, Marc Pollefeys, and Stefan Roth.
\newblock Semantic stixels: Depth is not enough.
\newblock In {\em IV}, 2016.

\bibitem{levi2015stixelnet}
Dan Levi, Noa Garnett, Ethan Fetaya, and Israel Herzlyia.
\newblock Stixelnet: A deep convolutional network for obstacle detection and
  road segmentation.
\newblock In {\em BMVC}, 2015.

\bibitem{benenson2012fast}
Rodrigo Benenson, Markus Mathias, Radu Timofte, and Luc Van~Gool.
\newblock Fast stixel computation for fast pedestrian detection.
\newblock In {\em ECCV}, 2012.

\bibitem{li2016new}
Xiaofei Li, Fabian Flohr, Yue Yang, Hui Xiong, Markus Braun, Shuyue Pan,
  Keqiang Li, and Dariu~M Gavrila.
\newblock A new benchmark for vision-based cyclist detection.
\newblock In {\em IV}, 2016.

\bibitem{franke2013making}
Uwe Franke, David Pfeiffer, Clemens Rabe, Carsten Knoeppel, Markus Enzweiler,
  Fridtjof Stein, and Ralf Herrtwich.
\newblock Making bertha see.
\newblock In {\em International Conference on Computer Vision Workshops}, pages
  214--221, 2013.

\bibitem{ziegler2014making}
Julius Ziegler, Philipp Bender, Markus Schreiber, Henning Lategahn, Tobias
  Strauss, Christoph Stiller, Thao Dang, Uwe Franke, Nils Appenrodt,
  Christoph~G Keller, et~al.
\newblock Making bertha drive—an autonomous journey on a historic route.
\newblock {\em Intelligent transportation systems magazine}, 6(2):8--20, 2014.

\bibitem{hehn2019instance}
Thomas~M Hehn, Julian~FP Kooij, and Dariu~M Gavrila.
\newblock Instance stixels: Segmenting and grouping stixels into objects.
\newblock In {\em IV}, 2019.

\bibitem{alhaija2017augmented}
Hassan~Abu Alhaija, Siva~Karthik Mustikovela, Lars Mescheder, Andreas Geiger,
  and Carsten Rother.
\newblock Augmented reality meets deep learning for car instance segmentation
  in urban scenes.
\newblock In {\em BMVC}, 2017.

\bibitem{liu2018path}
Shu Liu, Lu~Qi, Haifang Qin, Jianping Shi, and Jiaya Jia.
\newblock Path aggregation network for instance segmentation.
\newblock In {\em CVPR}, 2018.

\bibitem{neven2019instance}
Davy Neven, Bert~De Brabandere, Marc Proesmans, and Luc~Van Gool.
\newblock Instance segmentation by jointly optimizing spatial embeddings and
  clustering bandwidth.
\newblock In {\em CVPR}, 2019.

\bibitem{xiong2019upsnet}
Yuwen Xiong, Renjie Liao, Hengshuang Zhao, Rui Hu, Min Bai, Ersin Yumer, and
  Raquel Urtasun.
\newblock Upsnet: A unified panoptic segmentation network.
\newblock In {\em Proceedings of the IEEE Conference on Computer Vision and
  Pattern Recognition}, pages 8818--8826, 2019.

\bibitem{peng2020deep}
Sida Peng, Wen Jiang, Huaijin Pi, Xiuli Li, Hujun Bao, and Xiaowei Zhou.
\newblock Deep snake for real-time instance segmentation.
\newblock In {\em Proceedings of the IEEE/CVF Conference on Computer Vision and
  Pattern Recognition}, pages 8533--8542, 2020.

\bibitem{tatoglu2012point}
Akin Tatoglu and Kishore Pochiraju.
\newblock Point cloud segmentation with lidar reflection intensity behavior.
\newblock In {\em ICRA}, 2012.

\bibitem{tchapmi2017segcloud}
Lyne Tchapmi, Christopher Choy, Iro Armeni, JunYoung Gwak, and Silvio Savarese.
\newblock Segcloud: Semantic segmentation of 3d point clouds.
\newblock In {\em 3DV}, 2017.

\bibitem{le2018pointgrid}
Truc Le and Ye~Duan.
\newblock Pointgrid: A deep network for 3d shape understanding.
\newblock In {\em CVPR}, 2018.

\bibitem{wu20153d}
Zhirong Wu, Shuran Song, Aditya Khosla, Fisher Yu, Linguang Zhang, Xiaoou Tang,
  and Jianxiong Xiao.
\newblock 3d shapenets: A deep representation for volumetric shapes.
\newblock In {\em CVPR}, 2015.

\bibitem{maturana2015voxnet}
Daniel Maturana and Sebastian Scherer.
\newblock Voxnet: A 3d convolutional neural network for real-time object
  recognition.
\newblock In {\em IROS}, 2015.

\bibitem{huang2016point}
Jing Huang and Suya You.
\newblock Point cloud labeling using 3d convolutional neural network.
\newblock In {\em ICPR}, 2016.

\bibitem{qi2016volumetric}
Charles~R Qi, Hao Su, Matthias Nie{\ss}ner, Angela Dai, Mengyuan Yan, and
  Leonidas~J Guibas.
\newblock Volumetric and multi-view cnns for object classification on 3d data.
\newblock In {\em CVPR}, 2016.

\bibitem{riegler2017octnet}
Gernot Riegler, Ali Osman~Ulusoy, and Andreas Geiger.
\newblock Octnet: Learning deep 3d representations at high resolutions.
\newblock In {\em CVPR}, 2017.

\bibitem{dai2017scannet}
Angela Dai, Angel~X Chang, Manolis Savva, Maciej Halber, Thomas Funkhouser, and
  Matthias Nie{\ss}ner.
\newblock Scannet: Richly-annotated 3d reconstructions of indoor scenes.
\newblock In {\em CVPR}, 2017.

\bibitem{qi2017pointnet}
Charles~R Qi, Hao Su, Kaichun Mo, and Leonidas~J Guibas.
\newblock Pointnet: Deep learning on point sets for 3d classification and
  segmentation.
\newblock In {\em CVPR}, 2017.

\bibitem{qi2017pointnet++}
Charles~Ruizhongtai Qi, Li~Yi, Hao Su, and Leonidas~J Guibas.
\newblock Pointnet++: Deep hierarchical feature learning on point sets in a
  metric space.
\newblock In {\em NeurIPS}, 2017.

\bibitem{li2018pointcnn}
Yangyan Li, Rui Bu, Mingchao Sun, Wei Wu, Xinhan Di, and Baoquan Chen.
\newblock Pointcnn: Convolution on x-transformed points.
\newblock In {\em NeurIPS}, 2018.

\bibitem{jiang2018pointsift}
M~Jiang, Y~Wu, and C~PointSIFT Lu.
\newblock A sift-like network module for 3d point cloud semantic segmentation.
\newblock In {\em Comput. Vis. Pattern Recognit}, 2018.

\bibitem{he2017mask}
Kaiming He, Georgia Gkioxari, Piotr Doll{\'a}r, and Ross Girshick.
\newblock Mask r-cnn.
\newblock In {\em ICCV}, 2017.

\bibitem{yi2019gspn}
Li~Yi, Wang Zhao, He~Wang, Minhyuk Sung, and Leonidas~J Guibas.
\newblock Gspn: Generative shape proposal network for 3d instance segmentation
  in point cloud.
\newblock In {\em CVPR}, 2019.

\bibitem{qi2018frustum}
Charles~R Qi, Wei Liu, Chenxia Wu, Hao Su, and Leonidas~J Guibas.
\newblock Frustum pointnets for 3d object detection from rgb-d data.
\newblock In {\em CVPR}, 2018.

\bibitem{yang2019learning}
Bo~Yang, Jianan Wang, Ronald Clark, Qingyong Hu, Sen Wang, Andrew Markham, and
  Niki Trigoni.
\newblock Learning object bounding boxes for 3d instance segmentation on point
  clouds.
\newblock In {\em NeurIPS}, 2019.

\bibitem{cordts2016cityscapes}
Marius Cordts, Mohamed Omran, Sebastian Ramos, Timo Rehfeld, Markus Enzweiler,
  Rodrigo Benenson, Uwe Franke, Stefan Roth, and Bernt Schiele.
\newblock The cityscapes dataset for semantic urban scene understanding.
\newblock In {\em CVPR}, 2016.

\bibitem{liu2016ssd}
Wei Liu, Dragomir Anguelov, Dumitru Erhan, Christian Szegedy, Scott Reed,
  Cheng-Yang Fu, and Alexander~C Berg.
\newblock Ssd: Single shot multibox detector.
\newblock In {\em ECCV}, 2016.

\bibitem{huang2017speed}
Jonathan Huang, Vivek Rathod, Chen Sun, Menglong Zhu, Anoop Korattikara,
  Alireza Fathi, Ian Fischer, Zbigniew Wojna, Yang Song, Sergio Guadarrama,
  et~al.
\newblock Speed/accuracy trade-offs for modern convolutional object detectors.
\newblock In {\em CVPR}, pages 7310--7311, 2017.

\bibitem{murtagh1983survey}
Fionn Murtagh.
\newblock A survey of recent advances in hierarchical clustering algorithms.
\newblock {\em The computer journal}, 26(4):354--359, 1983.

\bibitem{dohan2015learning}
David Dohan, Brian Matejek, and Thomas Funkhouser.
\newblock Learning hierarchical semantic segmentations of lidar data.
\newblock In {\em 3DV}, 2015.

\end{thebibliography}


\begin{thebibliography}{99}

\bibitem{c1} G. O. Young, ÒSynthetic structure of industrial plastics (Book style with paper title and editor),Ó 	in Plastics, 2nd ed. vol. 3, J. Peters, Ed.  New York: McGraw-Hill, 1964, pp. 15Ð64.
\bibitem{c2} W.-K. Chen, Linear Networks and Systems (Book style).	Belmont, CA: Wadsworth, 1993, pp. 123Ð135.
\bibitem{c3} H. Poor, An Introduction to Signal Detection and Estimation.   New York: Springer-Verlag, 1985, ch. 4.
\bibitem{c4} B. Smith, ÒAn approach to graphs of linear forms (Unpublished work style),Ó unpublished.
\bibitem{c5} E. H. Miller, ÒA note on reflector arrays (Periodical styleÑAccepted for publication),Ó IEEE Trans. Antennas Propagat., to be publised.
\bibitem{c6} J. Wang, ÒFundamentals of erbium-doped fiber amplifiers arrays (Periodical styleÑSubmitted for publication),Ó IEEE J. Quantum Electron., submitted for publication.
\bibitem{c7} C. J. Kaufman, Rocky Mountain Research Lab., Boulder, CO, private communication, May 1995.
\bibitem{c8} Y. Yorozu, M. Hirano, K. Oka, and Y. Tagawa, ÒElectron spectroscopy studies on magneto-optical media and plastic substrate interfaces(Translation Journals style),Ó IEEE Transl. J. Magn.Jpn., vol. 2, Aug. 1987, pp. 740Ð741 [Dig. 9th Annu. Conf. Magnetics Japan, 1982, p. 301].
\bibitem{c9} M. Young, The Techincal Writers Handbook.  Mill Valley, CA: University Science, 1989.
\bibitem{c10} J. U. Duncombe, ÒInfrared navigationÑPart I: An assessment of feasibility (Periodical style),Ó IEEE Trans. Electron Devices, vol. ED-11, pp. 34Ð39, Jan. 1959.
\bibitem{c11} S. Chen, B. Mulgrew, and P. M. Grant, ÒA clustering technique for digital communications channel equalization using radial basis function networks,Ó IEEE Trans. Neural Networks, vol. 4, pp. 570Ð578, July 1993.
\bibitem{c12} R. W. Lucky, ÒAutomatic equalization for digital communication,Ó Bell Syst. Tech. J., vol. 44, no. 4, pp. 547Ð588, Apr. 1965.
\bibitem{c13} S. P. Bingulac, ÒOn the compatibility of adaptive controllers (Published Conference Proceedings style),Ó in Proc. 4th Annu. Allerton Conf. Circuits and Systems Theory, New York, 1994, pp. 8Ð16.
\bibitem{c14} G. R. Faulhaber, ÒDesign of service systems with priority reservation,Ó in Conf. Rec. 1995 IEEE Int. Conf. Communications, pp. 3Ð8.
\bibitem{c15} W. D. Doyle, ÒMagnetization reversal in films with biaxial anisotropy,Ó in 1987 Proc. INTERMAG Conf., pp. 2.2-1Ð2.2-6.
\bibitem{c16} G. W. Juette and L. E. Zeffanella, ÒRadio noise currents n short sections on bundle conductors (Presented Conference Paper style),Ó presented at the IEEE Summer power Meeting, Dallas, TX, June 22Ð27, 1990, Paper 90 SM 690-0 PWRS.
\bibitem{c17} J. G. Kreifeldt, ÒAn analysis of surface-detected EMG as an amplitude-modulated noise,Ó presented at the 1989 Int. Conf. Medicine and Biological Engineering, Chicago, IL.
\bibitem{c18} J. Williams, ÒNarrow-band analyzer (Thesis or Dissertation style),Ó Ph.D. dissertation, Dept. Elect. Eng., Harvard Univ., Cambridge, MA, 1993. 
\bibitem{c19} N. Kawasaki, ÒParametric study of thermal and chemical nonequilibrium nozzle flow,Ó M.S. thesis, Dept. Electron. Eng., Osaka Univ., Osaka, Japan, 1993.
\bibitem{c20} J. P. Wilkinson, ÒNonlinear resonant circuit devices (Patent style),Ó U.S. Patent 3 624 12, July 16, 1990. 

\end{thebibliography}

\end{document}